\pdfoutput=1
\documentclass{article} 
\usepackage{iclr2025_conference,times}

\usepackage{amsmath,amsfonts,bm}

\def\eqref#1{equation~\ref{#1}}

\def\1{\bm{1}}

\def\rvs{{\mathbf{s}}}

\def\rvv{{\mathbf{v}}}

\def\rvx{{\mathbf{x}}}

\def\rmB{{\mathbf{B}}}

\def\rmE{{\mathbf{E}}}

\def\rmL{{\mathbf{L}}}

\def\rmR{{\mathbf{R}}}
\def\rmS{{\mathbf{S}}}

\def\rmV{{\mathbf{V}}}
\def\rmW{{\mathbf{W}}}
\def\rmX{{\mathbf{X}}}

\DeclareMathAlphabet{\mathsfit}{\encodingdefault}{\sfdefault}{m}{sl}
\SetMathAlphabet{\mathsfit}{bold}{\encodingdefault}{\sfdefault}{bx}{n}

\usepackage{hyperref}
\usepackage{url}

\usepackage{dsfont}
\usepackage{algorithm}
\usepackage{algpseudocode}
\usepackage{amssymb}

\usepackage{mathtools}
\usepackage{multirow}
\usepackage{amsmath}
\usepackage{array}
\newcolumntype{?}{!{\vrule width 1pt}}
\usepackage{enumitem}
\usepackage{multirow}
\usepackage{wrapfig,booktabs}
\usepackage{booktabs} 

\usepackage{colortbl}
\usepackage{amssymb}
\usepackage{pifont}
\newcommand{\cmark}{\ding{51}}%
\newcommand{\xmark}{\ding{55}}%

\makeatletter
\newcommand{\thickhline}{%
    \noalign {\ifnum 0=`}\fi \hrule height 1.5pt
    \futurelet \reserved@a \@xhline
}
\newcolumntype{"}{@{\hskip\tabcolsep\vrule width 1pt\hskip\tabcolsep}}
\makeatother

\newcommand{\ours}{NoTS}
\newcommand{\ourslw}{NoTS-lw}

\title{Generalizable autoregressive modeling of time series through functional narratives}

\author{Ran Liu$^{1,2,*}$ Wenrui Ma$^{1,*}$ Ellen Zippi$^{2}$ Hadi Pouransari$^{2}$ Jingyun Xiao$^{1}$ Chris Sandino$^{2}$ \\ \textbf{Behrooz Mahasseni$^{2}$ Juri Minxha$^{2}$ Erdrin Azemi$^{2}$ Eva L. Dyer$^{1}$ Ali Moin$^{2}$} \\
  $^{1}$Georgia Tech, $^{2}$Apple, $^*$Equal contribution
}

\iclrfinalcopy 
\begin{document}

\maketitle

\begin{abstract}
  \vspace{-3mm}
  Time series data are inherently functions of time, yet current transformers often learn time series by modeling them as mere concatenations of time periods, overlooking their functional properties. In this work, we propose a novel objective for transformers that learn time series by re-interpreting them as temporal functions. We build an alternative sequence of time series by constructing degradation operators of different intensity in the functional space, creating augmented variants of the original sample that are abstracted or simplified to different degrees. Based on the new set of generated sequence, we train an autoregressive transformer that progressively recovers the original sample from the most simplified variant. Analogous to the next word prediction task in languages that learns narratives by connecting different words, our autoregressive transformer aims to learn the Narratives of Time Series (\ours) by connecting different functions in time. Theoretically, we justify the construction of the alternative sequence through its advantages in approximating functions. When learning time series data with transformers, constructing sequences of temporal functions allows for a broader class of approximable functions (e.g., differentiation) compared to sequences of time periods, leading to a 26$\%$ performance improvement in synthetic feature regression experiments. Experimentally, we validate \ours~in 3 different tasks across 22 real-world datasets, where we show that \ours~significantly outperforms other pre-training methods by up to 6\%. Additionally, combining \ours~on top of existing transformer architectures can consistently boost the performance. Our results demonstrate the potential of \ours~as a general-purpose dynamic learner, offering a viable alternative for developing foundation models for time series analysis.
\end{abstract}

\vspace{-10mm}
\begin{figure}[b!]
  \centering
  \includegraphics[width=0.9\textwidth]{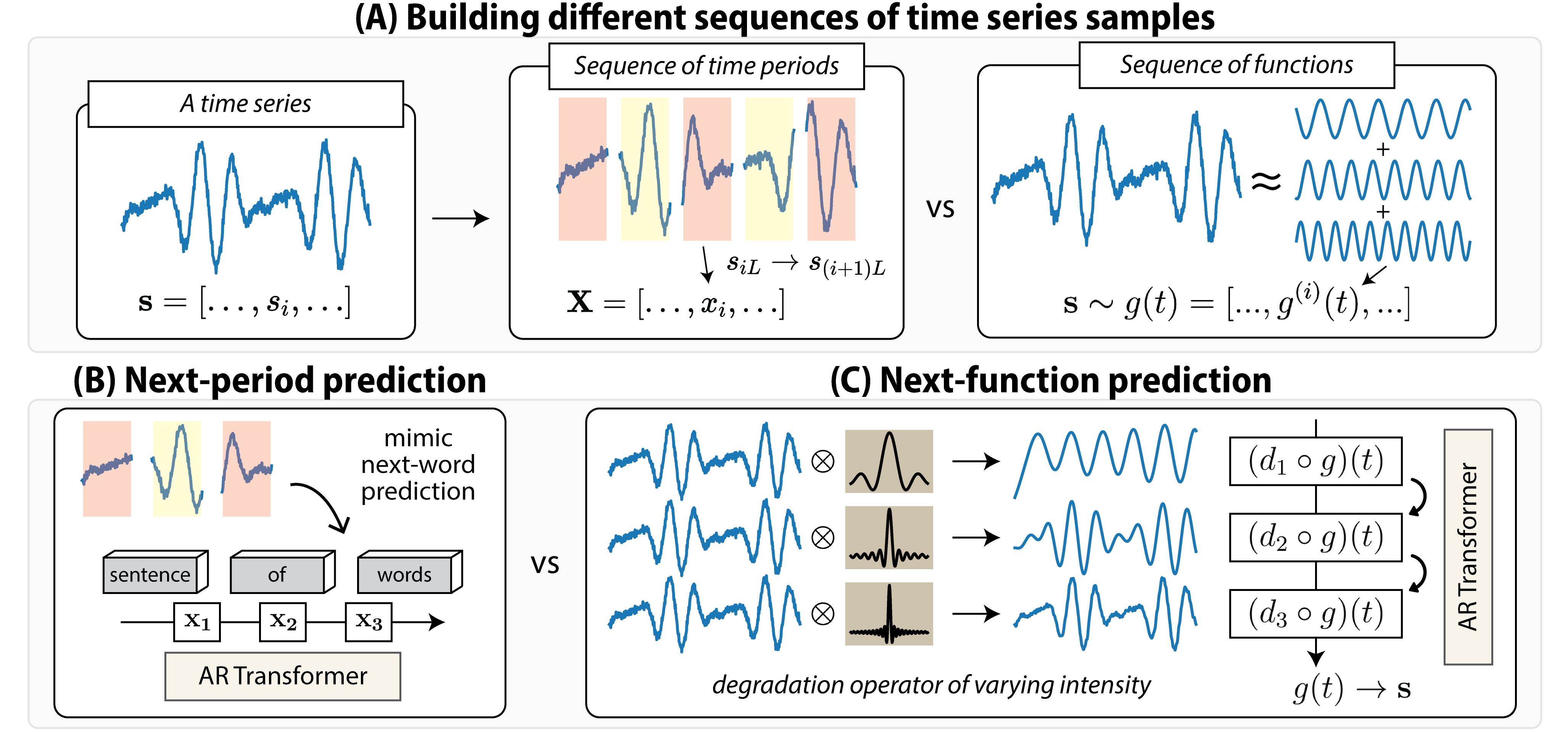}
  \vspace{-4mm}
  \caption{\footnotesize{\textit{Overview.} (A) Given a sample of time series, one can build different sequences from the original sample by treating it as either concatenation of time periods, or composition of temporal functions.
  (B) In the former case, it is common to emulate the next word prediction task in language to predict the next time period with an autoregressive (AR) transformer.
  (C) Alternatively, by applying degradation operators of varying intensity, we can craft augmented variants of samples that are progressively simplified, allowing a next-function prediction task. 
  The AR transformer is trained on the alternative sequence to learn the relationship across the sequence of functions to gradually recover the variance within original samples.}}
  \label{fig:overview}
\end{figure}

\section{Introduction}
\vspace{-2mm}

Recent advances in large language models (LLMs)
demonstrate the advantage of large-scale pre-training, providing a generalizable way for modeling complex systems \citep{radford2019language}.
At the core of state-of-the-art LLMs is the next-token prediction task \citep{achiam2023gpt}, where each data sample (sentence) is segmented into tokens (words), and the next word is predicted based on the previous words using the transformer architecture 
\citep{vaswani2017attention}.
By completing samples of sentences based on partial information in an autoregressive (AR) way, LLMs build generalizable data representations that can be rapidly adapted to new datasets and tasks \citep{brown2020language}.

Many works in time series analysis mimic the transformer-based modeling approaches in language by building sequences from samples through segmenting time series into periods of time points (Figure~\ref{fig:overview}(A) \citep{nie2022time,liu2022seeing,zhang2023crossformer}. An AR transformer on top of it would predict the next time period based on the existing ones (Figure~\ref{fig:overview}(B) \citep{garza2023timegpt}. 
However, this modeling approach
has two issues: (1) slicing time series into periods \textit{\textbf{breaks nonlocal functional properties}} like trend or periodicity, and often requires special remedies to compensate for the issues \citep{zhou2022fedformer}; (2) the predicted time periods \textit{\textbf{lack generalizability}}, as the prediction is sensitive to the length of chunks, the position of where the slicing happens, and the characteristics of datasets.
To compensate for the issues of patching and build generalizability into the transformer,
recent works rely on the usage of operators like Fourier neural operators or Koopman operators, but they either require specially engineered coding blocks \citep{liu2023frequency}, or a specific set of predetermined bases that 
may vary across datasets \citep{liu2024koopa}.


Inspired by \cite{tian2024visual} that replaces next-patch prediction with a next-resolution prediction task in computer vision,
in this work,
we re-think alternative approaches to build a coarse-to-fine sequence of time series by considering them as functions of time.
Instead of slicing time series into periods, we consider time series samples $\rmS$ as a sampled version of an underlying function $g(t)$ that 
can be structurally simplified in its functional form (Figure~\ref{fig:overview}(A)). 
Instead of mapping the sample onto fixed sets of basis like Taylor or Fourier series, we isolate functional components in a data-dependent way by building degradation operators $d_k(\cdot)$ of different intensity levels $k$ and progressively applying them on the signals.
By doing so, we generate an alternative sequence of samples consisting of augmented variants of the signal with increasing amount of information, offering an interconnected yet simplified representation of the original signal.
We train an autoregressive transformer to learn the connection of the different set of functionals, building a knowledge map of different functional components \footnote{For example, in brain decoding tasks, signals often contain cross-frequency coupling where low-frequency components drive the high-frequency components \citep{klimesch2018frequency,donoghue2020parameterizing}.}.
Analogous to the next word prediction task that learns narrative in languages by completing sentences, we denote our method as the Narratives of Time Series (NoTS) because it learns the functional narrative of temporal signals (Figure~\ref{fig:overview}(C)).

We first justify the construction of the alternative sequence using an intuitive function approximation analysis in Section~\ref{sec:theorysection}.
When learning time series with transformers, under the universal approximation framework \citep{yun2019transformers},
we show that learning time series as sequences of periods of time points would cause approximation issues, as performing sampling operation on commonly encountered time series signal processing operators (e.g., differentiation) 
creates discontinuous sequence-to-sequence functions.
Instead, such limitations can be bypassed by forming and learning from the alternative function sequences, 
as long as either (1) the constructed sequence is expressive, or (2) an expressive tokenizer \citep{ismailov2023three} is used before learning with transformers.
The analytical result is validated experimentally through a feature regression task on synthetic datasets. We show that \ours~significantly outperforms other pre-training methods when approximating features with real-world characteristics, showing its 
superior expressiveness both theoretically and experimentally.

We further validate \ours~in real-world time series datasets in a multitask setting, where we consider performance across 22 real-world datasets consisting of classification, imputation, and anomaly detection tasks.
Across the board, \ours~improves the average performance of other pre-training methods by up to $6\%$,
significantly outperforming the state-of-the-arts including next-period prediction \citep{garza2023timegpt}, masked autoencoder (MAE) \citep{dong2024simmtm}, and MAE with Fourier neural operators \citep{liu2023frequency}.
Moreover, we show that \ours~can improve the performance of existing transformer architectures \citep{nie2022time,liu2023itransformer}, giving a consistent performance boost when performing dataset-specific pre-training. Interestingly, we present a synthetically pre-trained lightweight model \ourslw,
which can be efficiently adapted to real-world tasks and achieve $82\%$ average performance with only \textless $1\%$ parameters trained,
showing the potential 
of \ours~on learning dynamics that can be transferred across datasets and tasks.

The main contributions of this paper are summarized as follows:
\vspace{-2mm}
\begin{itemize}[noitemsep,leftmargin=*]
    \item An alternative approach to form sequences from time series data by considering them as functions of time and isolating functional components with constructed degradation operators.
    \item Analytical results showing that learning time series from the functional perspective allows the approximation of a broader class of functions when compared to learning across periods of time.
    \item A novel transformer-based pre-training framework
    \ours~that progressively reconstruct time series from their degraded variants, and thus learn the interrelationships across functions.
    \item Experimental results on 2 synthetic and 22 real-world datasets, including 4 different classes of tasks (classification, regression, anomaly detection, and imputation), showing that \ours~significantly outperforms other pre-training methods from next-period predictors to Fourier-informed masked autoencoders, giving a stable performance boost on top of existing architectures. 
    \item A synthetically pre-trained lightweight model \ourslw~that can be efficiently adapted on new datasets and tasks with \textless $1\%$ parameters trained while maintaining $82\%$ average performance.
\end{itemize}

\vspace{-2mm}
\section{Preliminaries and related works}
\vspace{-2mm}
\subsection{Preliminaries}
\vspace{-2mm}

\paragraph{Autoregressive (AR) transformers} 
AR transformers have revolutionized natural language processing by building next-token prediction-based language models \citep{ray2023chatgpt}. The transformer architectures learn the interactions across different elements (tokens) in a sequence $\rmX = [\rvx_1, \rvx_2, ..., \rvx_N]$,
and the AR objective is defined as follows: The probability of obtaining the next token $\rvx_i$ can be deduced from the observed subsequence $[\rvx_1, \rvx_2, ..., \rvx_{i-1}]$. Thus, the probability of obtaining the whole sample is the product of a sequence of unidirectional conditional probabilities:
\vspace{-2mm}
\begin{equation}
\label{eq:ar_basic}
p\left(\rvx_1, \rvx_2, \ldots, \rvx_N\right)=\prod_{i=1}^N p\left(\rvx_i \mid \rvx_1, \rvx_2, \ldots, \rvx_{i-1}\right)
\end{equation}
where the AR relationship is learned by a transformer model $p_\theta$ parameterized by $\theta$. In the language domain, $\rvx_i \in \mathcal{V}$ is typically a discrete token of a word from a given vocabulary, which forms next-word predictors with impressive in-context generalization capabilities \citep{brown2020language}.


\vspace{-1mm}
\paragraph{Notations for time series} Time series samples are sequences of data points from multiple channels.
A multivariate time series sample of $C$ channels and a length of $T$ is represented as 
$\rmS = [\rvv_1, \rvv_2, ..., \rvv_T] \in \mathbb{R}^{C \times T}$. Typically, each dataset has its unique channel-wise relationships.

To apply transformers on time series, one needs to form sequences from the given signal $\rmS$. A naive approach  is to directly treat $\rvv_i$ as tokens and then apply transformers \citep{zerveas2021transformer}. The drawback is that the token representation space is dependent only on $\rvv_i$, which varies across datasets, making it less generalizable. Recently, many time series framework produce tokens through cutting time series into different periods of time with a length of $L$, which creates tokens $\rvx_i = \operatorname{Tokenizer}([\rvv_{iL}, ..., \rvv_{(i+1)L}])$ that contain more dynamics \citep{ren2022autotransformer}. 
To further eliminate the negative impact of channel-wise relationships on generalizability, 
\cite{nie2022time,liu2022seeing} considers the channel-independent design, which processes each channel (row) of $\rmS$ independently, producing tokens based on individual channels for transformers. While the approach demonstrated more generalizability, it is computationally expensive in high-density settings, which was later discussed by other works \citep{zhang2023crossformer}.

\vspace{-2mm}
\subsection{Pretraining methods for time series}
\vspace{-2mm}

Pre-training on large-scale datasets has proven effective in helping models learn generalizable patterns, which is particularly advantageous in the time series domain where downstream datasets are often small-scale
\citep{liu2021drop,zhang2022self,woo2022cost}. There are two prominent approaches for reconstruction-based pre-training in transformers: masked modeling and next-period prediction.
Masked modeling trains transformers by randomly masking elements, and predicting the masked values with the remaining sequence. Representative works include SimMTM \citep{dong2024simmtm}, which implements masked modeling through aggregating neighboring points, and bioFAME \citep{liu2023frequency}, which employs Fourier-based kernels to achieve the same objective.
However, these methods suffer from the loss of nonlocal information. Several approaches are proposed to mitigate this issue, such as leveraging multi-resolution patches \citep{das2023decoder,woo2024unified}.
More recently, the next-period prediction approach has gained attention for pre-training, particularly in the development of foundation models. 
For instance, Time-GPT1 \citep{garza2023timegpt} implements next-period prediction in a straightforward manner, while Chronos \citep{ansari2024chronos} applies scaling and quantization techniques to tokenize time series data and model the categorical distributions. 
Despite their success, these models also suffer from the challenge of losing nonlocal information, which is partially addressed through the use of lagged features and temporal covariates in Lag-Llama \citep{rasul2023lag}.
Based upon previous works, our work aims to fundamentally address the issue through building and learning sequences from the function perspective.

An alternative line of research seeks to adapt language models directly for time series applications.
Some approaches transfer pre-trained weights from language models and retrain the tokenization layers to handle time-series-specific tasks \citep{zhou2023one,cao2023tempo,liu2024autotimes}. 
Another line of work focuses on reprogramming time series data into text, and applying LLMs to process the textual inputs with time series prompts \citep{jin2023time,xue2023promptcast}. 
While these works primarily aim to bridge the modality gap between pre-trained LLMs and time series applications, our approach is specifically designed for time series data to capture the subtle variations and dynamics inherent in temporal signals.



\vspace{-2mm}
\section{Methods}
\vspace{-2mm}

Alternative to modeling time series as sequences of fragmented time periods, our framework is built on the idea to model time series as sequences of constructed temporal functions with transformers.
We begin by introducing the high-level objective in Section~\ref{subsec:method1}, and then introduce the pre-training method \ours~in Section~\ref{subsec:method2} as well as how to adapt it in real-world tasks in Section~\ref{sec:prompt-tuning-method}.

\vspace{-2mm}
\subsection{The next-function prediction task}
\label{subsec:method1}

We assume that each signal $\rmS = [\rvv_1, \rvv_2, ..., \rvv_T] \in \mathbb{R}^{C \times T}$ is intrinsically controlled by a temporal function $g(t): \mathbb{R} \to \mathbb{R}^{C}$, where $\rvv_i = g(i)$ is the product of a sampling process in time. 
To train a transformer with awareness of the functional perspective, we build sequences of functions, where each element is a simplified version of the original sample.
Practically, we ask two questions: 
\begin{itemize}[noitemsep,leftmargin=*]
    \item How to build meaningful functional elements, as they tend to change across different datasets?
    \item How to form a meaningful sequence for the transformer, so that we can enforce the transformer to learn generalizable representation from the constructed sequences of functional components?
\end{itemize}

In this work, we propose to construct \emph{degradation functions} $d_k(\cdot)$ of intensity $k$, that generate augmentations of the original function $g(t)$ with varying levels of partial information. Applying degradation functions $d_k(\cdot)$ on signals generates data-dependent functions as tokens for transformer, removing the need for a fixed set of bases. By controlling the intensity of degradation, we create $g_k(t) = (d_k \circ g)(t)$, where $g_{k+1}(t)$ contains strictly more or an equal amount of information than $g_{k}(t)$ about the original sample $g(t)$, establishing a progressive relationship for the transformer to learn. 
Based on the constructed sequence, the new modeling approach becomes the following:
\vspace{-1mm}
\begin{equation}
\label{eq:ours_main_overview}
\begin{aligned}
& p\left(g_1(t), g_2(t), \ldots, g_K(t)\right)=\prod_{k=1}^K p\left(g_k(t) \mid g_1(t), g_2(t), \ldots, g_{k-1}(t)\right),
\end{aligned}
\end{equation}
where $K \rightarrow \infty$, $\rvv_i = g_{\infty}(i)$ forms the actual time series $\rmS$ under the sampling operation.

\begin{figure}[t!]
  \centering
  \includegraphics[width=\textwidth]{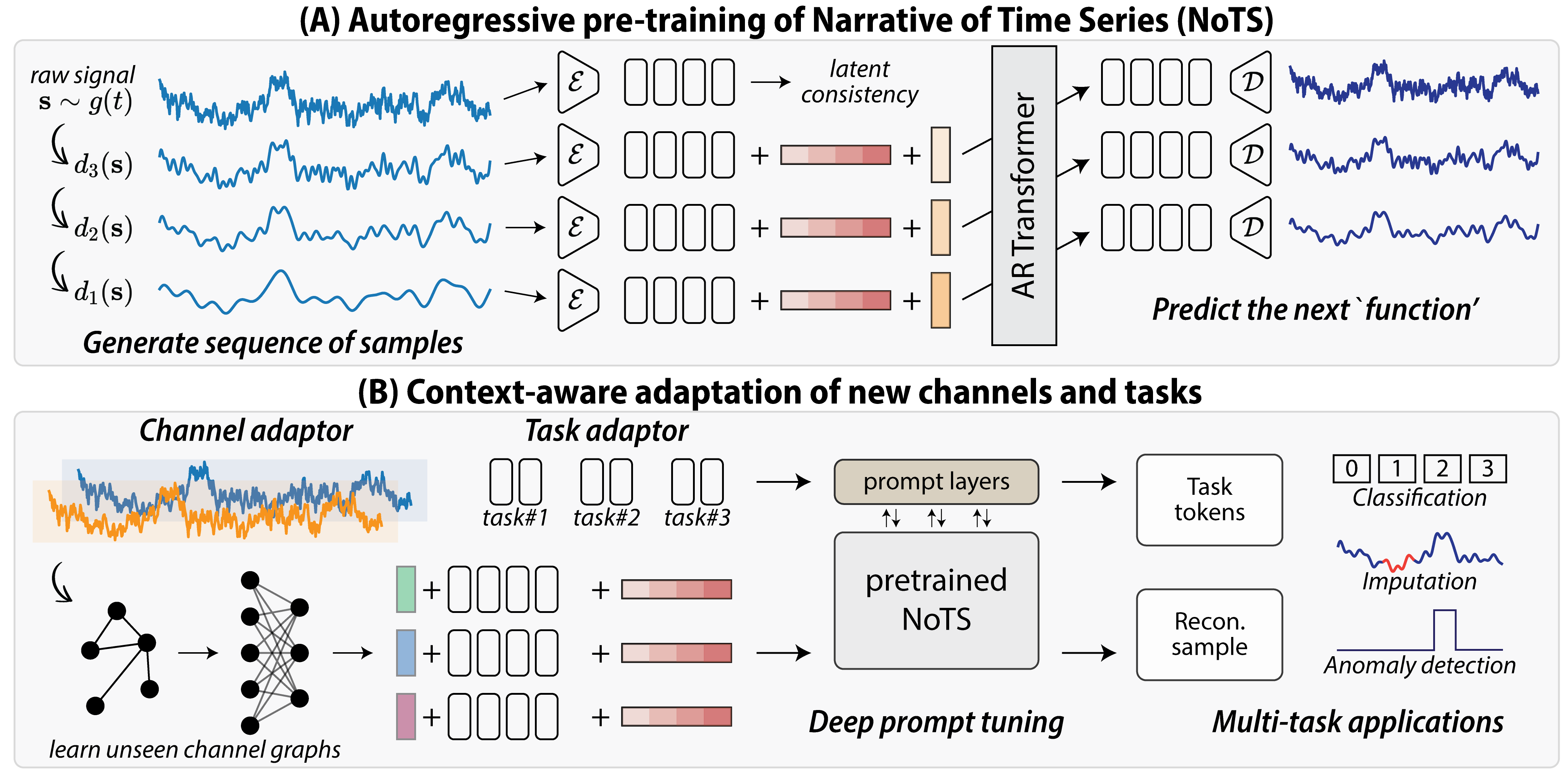}
  \caption{\footnotesize \textit{Narrative of Time Series (\ours)}. (A) To perform autoregressive pre-training of \ours, we first generate a sequence of time series from the raw signal that progressively simplifies the sample. The generated signals are passed into an encoder, added with position and resolution embeddings before fed into the AR transformer, which is trained with a decoder to reconstruct the signal of the next resolution. The raw signal was passed into a latent consistency loss directly. (B) To apply a pre-trained model on real-world dataset, we construct channel adaptor and task adaptor that handles unseen channel graphs and new tasks, respectively. The channel adaptor consists of a multi layer perceptron that pre-process channel maps and new additive channel embeddings. The task adaptor is newly initialized tokens that are prompted into the transformer following \cite{jia2022visual}. The produced task tokens and reconstructed samples are later used in multitask applications through this context-aware adaptation pipeline. }
\end{figure}

\vspace{-1mm}
\subsection{\ours: A novel pre-training objective for transformers}
\label{subsec:method2}

Based on the framework, the pre-training objective \ours~consists of the following components.
\vspace{-1mm}
\paragraph{Local and global degradation functions} Practically, we construct degraded signals with convolution operators: $\rmS_k = d_k(\rmS) =  (\rmS * w_k)[n]$,
where $*$ represents discrete convolution between rows of signal $\rmS$ and a kernel $w_k$. We use two different kernels as defined below:
\begin{itemize}[noitemsep,leftmargin=*]
    \item \textit{Local smoothing:} A simple averaging kernel of different lengths $p_k$ is used for local smoothing. Specifically, $w_k[n] = 1/p_k$ for $-0.5p_k \leq n \leq 0.5p_k$ and $w_k[n] = 0$ elsewhere. The set of even numbers $\{p_k\}$ is selected as hyperparameters with descending order as $k$ increases.
    \item \textit{Global smoothing:} A low-pass filter with different frequency cutoff values is used for global smoothing. Specifically, we build $w_k[n] = \operatorname{sinc}(p_k n)$, where $\{p_k\}$ is a set of values that control the frequency cutoff of $0.5p_k$ as hyperparameters with descending order as $k$ increases.
\end{itemize}

By constructing both local and global smoothing degradation functions, the proposed method can simultaneously model autoregressive relationships across different smoothness and frequencies, covering a prevalent range of tasks in both time and frequency. 

\paragraph{Autoregressive modeling of groups of tokens in the latent space}
We build tokenizers to convert the constructed signals $\rmS_k$ into recognizable embeddings for transformers. Ideally, we want to use one token for each function, yet this is computationally infeasible as the length of signals increases. Instead, we rely on existing encoder/decoder architectures as tokenizers to convert signals into group of tokens in the latent space to perform AR modeling. 
Specifically, the encoder produces groups of tokens from each signal $\rmR_k = \mathcal{E}(\rmS_k)$ and the decoder reconstructs signals from groups of tokens $\rmS_k^{\prime} = \mathcal{D}(\rmR_k^{\prime})$,
where each $\rmR_k$ and $\rmR_k^{\prime}$ consists of multiple tokens for the transformer.
The AR modeling is enforced by applying group-wise masking to the transformer attention map, where:
\[
[\rmR_{2}^{\prime}, ..., \rmR_{K}^{\prime}] = \operatorname{Transformer}([\rmR_{1}, ..., \rmR_{K-1}]), 
\text{ with } 
\operatorname{mask}[\Omega_k] = \begin{cases} 
      0, & \bigcup_{m=1}^{k} \Omega_{m} 
      \\
      -\infty, & \text{elsewhere}
      \end{cases}
\]
and $\Omega_k$ represents the set of sequence positions corresponding to $\rmR_k$.
While $\mathcal{E}$ and $\mathcal{D}$ can be any encoder/decoder architectures, we implement a lightweight model \ourslw~with a simple channel-independent 1D-ResNet encoder/decoder block to maintain fidelity in the token space. We also report results with different encoder/decoder architectures in Table~\ref{tab:main}.

\paragraph{Positional embeddings}
Since transformers are inherently invariant to the order of sequences, it is critical to embed sufficient information about the raw position into the transformer to ensure $\rmR_k$ includes sufficient information about samples. Thus, we add the following embeddings:

\begin{itemize}[noitemsep,leftmargin=*]
    \item \textit{Group embeddings}. To help transformers learn sufficient information about each function embedded by $\rmR_k$, we apply rotary positional embedding \citep{su2024roformer} on each group of tokens to encode the relative sequence position within each group or cohort of tokens.
    \item \textit{Degradation embeddings}. We add learnable absolute positional embeddings that encode the relative degree of degradation of each augmented variant of signals. In other words, the degradation embeddings encode information $k$ of the degradation function $d_k(\cdot)$.
    \item \textit{Channel embeddings (optional)}. When applying channel-independent architectures as the encoder, we add an additional set of learnable absolute positional embeddings along with the group embeddings to help encode channel-wise relationship within each group of tokens.
\end{itemize}

\paragraph{Training objective} 
We perform a self-supervised autoregressive reconstruction task to learn meaningful representations of time series using the proposed framework. 
To achieve this, we minimize the differences between $\rmS_k$ and the reconstructed $\rmS_{k+1}^{\prime}$ for every $k<K$. 
If only the AR loss is considered, the latent of the raw data $\rmR_K$ would remain unused by encoder/decoder throughout the training process. To avoid the resulting distributional shift, we add a \textit{latent consistency} term, that regularizes the consistency between the latent of the raw data to be able to be reconstructed back.
Thus, 
for each input signal matrix $\rmS \in \mathbb{R}^{C \times T}$, 
the training optimizes the following loss:
\begin{equation}
    \mathcal{L} = \sum_{k=1}^{K-1} \underbrace{\mathcal{L}_{\text{recon}}(\rmS_{k+1}^{\prime}, \rmS_k)}_{\text{AR reconstruction}} + \underbrace{\mathcal{L}_{\text{recon}}(\mathcal{D}(\mathcal{E}(\rmS_K)), \rmS_K)}_{\text{latent consistency term}}
\end{equation}
where $\mathcal{L}_{\text{recon}}$ is the reconstruction loss (mean absolute error is used throughout the paper).

\subsection{Context-aware adaptation through channel and task adaptors}
\label{sec:prompt-tuning-method}

In time series domain, it is common to encounter datasets with new channel graphs or new tasks at test time. To apply the pre-trained weights in such situations, we build two types of adaptors:



\paragraph{Channel adaptors} To learn new channel-wise relationship at test time, we build channel adaptors as follows: 
(1) To encourage information exchange across channels, we add a data embedding layer before applying the encoder $\mathcal{E}$. The data embedding layer is a simple linear layer that encodes on the channel dimension $\mathbb{R}^{C} \rightarrow \mathbb{R}^{C^{\prime}}$ to mix channel-wise information at an early stage. (2) We also re-initialize and re-train additive channel tokens for each dataset when applicable.

\paragraph{Task adaptors} To apply the pre-trained models on a diverse set of tasks, we build task adaptors as follows:
(1) We initialize and append prompt tokens to the transformer architecture along with data tokens following the deep visual prompt tuning plan as detailed in \cite{jia2022visual}; (2) We also add task-specific linear layers at the end of the transformer for inference purpose.

Given the two adaptors, we can perform context-aware adaptation of \ours~on new channel maps and tasks, allowing the transfer of general-purpose dynamics that are learned at the pre-training stage. Interestingly, the adaptation pipeline is also parameter-efficient: The adaptors add \textless $1\%$ new parameters in comparison to the original model consists of encoder, transformer, and decoder.

\section{An intuitive example: Approximating functions}
\label{sec:theorysection}

To justify the construction of functional sequences, we provide an intuitive example to investigate the expressive power of transformers under the context of time series domain. Following \cite{yun2019transformers,luo2022your}, we consider the standard transformer architectures:
\[
\mathcal{T}^{h, m, r}:=\left\{f: \mathbb{R}^{d \times n} \rightarrow \mathbb{R}^{d \times n} \mid f \text { consists of Transformer blocks } t^{h, m, r} \right\} .
\]
where $t^{h, m, r}$ consists of one self-attention layer of $h$ heads of size $m$ and one feed-forward layer with $r$ hidden dimensions (see definitions in Appendix Eq.~\ref{app:attention_basics}). To remove the restriction of permutation equivariance, we consider adding absolute positional embedding\footnote{Refer to \cite{luo2022your} for a case study of relative positional embedding under the UA framework.} 
to the transformer that creates $\mathcal{T}_{\mathrm{P}}^{h, m, r}:=\{f_{\mathrm{P}}(\boldsymbol{X})=f(\boldsymbol{X}+\boldsymbol{E})\}$ where $f \in \mathcal{T}^{h, m, r}$ and $\boldsymbol{X}, \boldsymbol{E} \in \mathbb{R}^{d \times n}$. Detailed in Appendix~\ref{app:theorysec}, our analysis is an extension of previous results in \cite{yun2019transformers,ismailov2023three}.

\subsection{Time series in the function space}

This work assumes time series data as functions in time $g(t)$, which forms function space $\mathcal{F}_{g}(\mathbb{R})$. An operator on the function space maps the original function $g(t)$ to a target function $h(t)$, creating a mapping across function spaces $A: \mathcal{F}_{g}(\mathbb{R}) \rightarrow \mathcal{F}_{h}(\mathbb{R})$. 
Assume that the signal is produced with a sampling plan $\{t_i\}_{i=1}^T$, the sampling operation on top of the functions discretizes the mapping into a set of output $\{A[g(t_i)]\}_{i=1}^{T}$, which forms a sequence-to-sequence function $f_{(A)}:\mathbb{R}^{d \times T} \rightarrow \mathbb{R}^{d \times T}$.
When breaking time series into concatenations of time periods, $\rmS$ is directly treated as inputs to the transformer $\rmX$, where one aims to find a transformer network $f_P \in \mathcal{T}_P^{h, m, r}$ to approximate $f_{(A)}$.

\paragraph{Example: The differential operator} It is intuitive that one can easily construct a linear but discontinuous mapping $A$, which is not necessarily approximable by transformers. See below:

\textbf{Theorem 1.} Given $T > 2$, and $\mathcal{D} \subseteq \mathbb{R}^{d \times T}$. Consider the differential operator $A$ that forms a sequence-to-sequence function $f_{(A)}$ under sampling plans $\{t_i\}_{i=1}^T$ with its initial starting point $t_1 \in \mathbb{R}$ and fixed intervals. There exists a $\rmX \in \mathbb{R}^{d \times T}$, such that:
\begin{equation}
\label{eq:prop1}
    \sup _{\boldsymbol{X} \in \mathcal{D}}\left\|f_P(\boldsymbol{X})-f_{(A)}(\boldsymbol{X})\right\|_2^2 \geq T
\end{equation}
holds for any transformer network $f_P \in \mathcal{T}_P^{h, m, r}$.

\textbf{Proof.} We construct a negative example with $d=1$. Consider a set of input functions $g_M(t) = \sin(Mt)/M$, the target functions under the differential operator are $h_M(t) = \cos(Mt)$. As $M$ increases, the input function converges uniformly to a constant zero function, which gives a sampled input matrix $\boldsymbol{X} \rightarrow \mathbf{0} \in \mathcal{D}$. 
At limit, the studied transformer network $f_P(\boldsymbol{X})$ converges to a fixed matrix $\rmB$ (see Appendix~\ref{app:lemma1}). Thus, given a sampling plan of interval $t_{i+1} - t_i = \pi/M$ and two initial starting points $t_1^{(1)}=0$ and $t_1^{(2)}=\pi$, we form $\boldsymbol{X}_1$ and $\boldsymbol{X}_2$ that give:
\vspace{-2mm}
\begin{equation}
\lim_{M \rightarrow \infty} \left\|f_P(\boldsymbol{X}_1)-f_{(A,M)}(\boldsymbol{X}_1)\right\| + \left\|f_P(\boldsymbol{X}_2)-f_{(A,M)}(\boldsymbol{X}_2)\right\| \geq \sum_{i=1}^{T} 2 = 2T 
\end{equation}
where $f_{(A,M)}$ denotes the function formed from $\{A[g_M(t_i)]\}_{i=1}^{T}$, which leads to Eq.~\ref{eq:prop1}.

\subsection{Two sufficient conditions to approximate the differential operator}

When considering signals as functions in time, sampling from simple signal processing operators may create discontinuous sequence-to-sequence functions, causing approximation issues of transformer if one directly considers $\rmS$ as inputs $\rmX$ to the transformer. Instead, by constructing signals sequences of length $T$ using $\rmS_k = d_k(\rmS)$, and performing dimensionality reduction with an encoder $\mathcal{E}$, we create two sufficient conditions to address the approximation issue as follows:

\textbf{Proposition 1.} Given a signal $\rmS \in \mathbb{R}^{d \times T}$ and an encoder $\mathcal{E}: \mathbb{R}^{d \times T} \rightarrow \mathbb{R}^{d}$, there exists two sufficient conditions to approximate $\{A[g(t_i)]\}_{i=1}^{T}$ with the construction of $\rmX = [\mathcal{E}(\rmS_1), \mathcal{E}(\rmS_2), ..., \mathcal{E}(\rmS_T)]$:
\begin{itemize}[noitemsep,leftmargin=*,topsep=0pt]
    \item The constructed $\rmS_i$ is expressive such that there exists a continuous mapping between a fixed element of $\rmS_i$ and the $i$-th element of the target output $A[g(t_i)]$;
    \item Given any distinguishable $\rmS_i$, there exists an expressive tokenizer $\mathcal{E}$ that preprocess $\rmS_i$ to create a continuous mapping between $\mathcal{E}(\rmS_i)$ to the target.
\end{itemize}
\textbf{Proof.} See Appendix~\ref{app:proposition}.

\paragraph{Example solutions} We provide example solutions to approximate the differential operator. Under the first condition, a trivial solution can be constructed by performing phase transition with degradation operator, creating $(d_i \circ g)(t) = \sin(Mt+\frac{i\pi}{2T})/M$. While the first condition requires data-specific degradation operators, 
the second condition provides more flexibility. In the case of differential operator, we rely on the result from \cite{ismailov2023three} and use a Kolmogorov’s mapping three-layer neural network to approximate an arbitrary (continuous or discontinuous) function $f: \mathbb{I}^T \to \mathbb{R}$, where $\mathbb{I}$ is a compact interval $[0, 1]$.
Thus, one can construct a simple degradation operator of value shifts as:
$(d_i \circ g)(t) = \sin(Mt)/M + (1+i\delta/T)/M$,
where $\delta \in (0, M-2]$ is an arbitrary number that distinguishes $\rmS_i$. 
In this case, there exists an encoder $\mathcal{E}$ that can create a continuous sequence-to-sequence function to the desired target (e.g., $\mathcal{E}(\rmS_i) = Mt_i$),
where the existence of a solution is guaranteed by previous results in \cite{yun2019transformers}.


\section{Experimental Results}


While the approximation analysis 
posts strong assumptions on the solution including the minimal length $T$ of the constructed sequence and the use of specific encoder $\mathcal{E}$. In this section, we show that, experimentally, \ours~works in both synthetic and real-world applications with relaxed assumptions.

 \begin{table}[t!]
  \begin{center}
    \caption{\footnotesize Feature approximation results on synthetic datasets. We compare the function approximation ability of different pre-training methods given the same architecture and pre-training pipeline. All presented numbers are averaged across three runs and scaled by 100 for better readability. Lower numbers are better.
    }
    \label{tab:synthetic}
    \vspace{-2mm}
 \resizebox{\linewidth}{!}{%
      \begin{tabular}{c|ccc|ccc} 
       & \multicolumn{3}{c|}{\textit{Fractional Brownian motion (fBm)}} & \multicolumn{3}{c}{\textit{Autocorrelated sinusoids}} \\
     Regression ($\downarrow$) & $\mathcal{H}$-index (1D) & SSC (32D) & WAMP (32D) & SSC (32D) & WAMP (32D) & b. power (96D) \\
       \hline
{VQVAE}  & 3.78 $\pm$ \footnotesize{0.45} & 38.93 $\pm$ \footnotesize{0.70} & 65.77 $\pm$ \footnotesize{3.72} & 26.24 $\pm$ \footnotesize{0.61} & 29.13 $\pm$ \footnotesize{0.90} & 14.37 $\pm$ \footnotesize{0.03} \\
{MAE} & 2.01 $\pm$ \footnotesize{0.61} & 25.78 $\pm$ \footnotesize{0.11} & 26.34 $\pm$ \footnotesize{0.03} & 25.29 $\pm$ \footnotesize{0.31} & 28.81 $\pm$ \footnotesize{2.86} & 14.90 $\pm$ \footnotesize{0.02} \\
{FAMAE} & 1.99 $\pm$ \footnotesize{0.24} & 33.85 $\pm$ \footnotesize{0.53} & 45.76 $\pm$ \footnotesize{0.24} & 28.26 $\pm$ \footnotesize{0.57} & 24.82 $\pm$ \footnotesize{0.84} & 13.92 $\pm$ \footnotesize{0.02} \\
{Next-period pred.} & 1.75 $\pm$ \footnotesize{0.11} & 27.38 $\pm$ \footnotesize{0.12} & 26.66 $\pm$ \footnotesize{0.19} & 24.44 $\pm$ \footnotesize{0.11} & 28.97 $\pm$ \footnotesize{1.37} & 13.96 $\pm$ \footnotesize{0.04} \\
{\bf \ours~(Ours)} & 1.27 $\pm$ \footnotesize{0.16} & 23.78 $\pm$ \footnotesize{0.34} & 20.04 $\pm$ \footnotesize{0.12} & 23.13 $\pm$ \footnotesize{0.19} & 24.58 $\pm$ \footnotesize{0.48} & 13.62 $\pm$ \footnotesize{0.05} \\
       \hline
\cellcolor{cyan!10}Improvement & \cellcolor{cyan!10}\textbf{$\uparrow$ 37.80$\%$} & \cellcolor{cyan!10}\textbf{$\uparrow$ 8.41$\%$} & \cellcolor{cyan!10}\textbf{$\uparrow$ 31.44$\%$} & \cellcolor{cyan!10}\textbf{$\uparrow$ 5.66$\%$} & \cellcolor{cyan!10}\textbf{$\uparrow$ 0.98$\%$} & \cellcolor{cyan!10}\textbf{$\uparrow$ 2.20$\%$} \\
    \end{tabular}
 }
 \vspace{-5mm}
  \end{center}
\end{table}

\vspace{-2mm}
\subsection{Synthetic experiments: A feature regression task}
\vspace{-1mm}
\paragraph{Datasets} We build two synthetic datasets with AR components in time and frequency spaces.

\underline{\smash{(1) Fractional Brownian motion (fBm).}}
fBm is a generalized Gaussian process with special covariance structure that was found to be similar to many types of time series datasets such as traffic, stock prices, and biosignals \citep{rivero2016new}. Unlike the classic Brownian motion, fBm has interdependent increments across time that are controlled by the Hurst index $\mathcal{H} \in (0, 1)$, which creates autoregressive components in time that exhibit long-range ($\mathcal{H}>0.5$) or short-range ($\mathcal{H}<0.5$) dependencies.
We simulate the fBm process 20,000 times to create signals of length 1024 using the Cholesky's Method \citep{dieker2003spectral} with $O(l^3)$ complexity, and remove the generated signals with abnormal values due to simulation instability (around 0.5\% of all data).

\underline{\smash{(2) Superposition of autocorrelated sinusoids.}} We extend previous synthetic datasets based on sinusoids \citep{yoon2019time,das2023decoder}, and build a new synthetic dataset of sinusoids with AR components in the frequency space. 
Specifically, we sample the set of $\{f_i\}_{i=1}^{B}$ based on five random AR(B/2) processes, where we set $B=\{20, 16, 10, 8, 4\}$.
We generate amplitude following the $a_i = 1/f_i$ frequency distribution, uniformly sample the phase $p_i \in (0, 2\pi]$, and add Gaussian noise $0.05*\mathcal{N}(0, 1)$ to the signal.
We randomly initialize each process 10,000 times, creating a dataset of 50,000 samples where each sample is of length 1024. 



\paragraph{Results on feature regression} 
We estimate different pre-training methods' capability of approximating functions with the feature regression task. 
The ground truth of features are built based on common signal processing analysis methods Slope Sign Change (SSC, 32D) and Willison Amplitude (WAMP, 32D), and we also include the Hurst index ($\mathcal{H}$-index, 1D) for the fBM dataset and the band power (b. power, 96D) for the sinusoids (Appendix~\ref{app:synthetic_exp}).
Note that SSC and WAMP are both implemented with global thresholding, making them discontinuous sequence-to-sequence functions.
Following Section~\ref{sec:prompt-tuning-method}, we train a VQVAE \citep{van2017neural}, masked autoencoder (MAE) \citep{dong2024simmtm}, frequency-aware MAE (FAMAE) \citep{liu2023frequency}, next-period prediction transformer \citep{garza2023timegpt}, and \textbf{\ourslw~}on the synthetic datasets by appending them with a regression task adaptor to validate the performance of our proposed method.

As shown in Table~\ref{tab:synthetic}, across the board, \textbf{\ourslw}~significantly outperforms all other pre-training methods given the same architecture and training pipeline. The relative improvement is especially pronounced in the fBm dataset, where data has complicated covariance architecture that was found relevant in many real-world applications, where we have $26\%$ improvements across the features.  

\begin{figure}[t!]
  \centering
  \includegraphics[width=\textwidth]{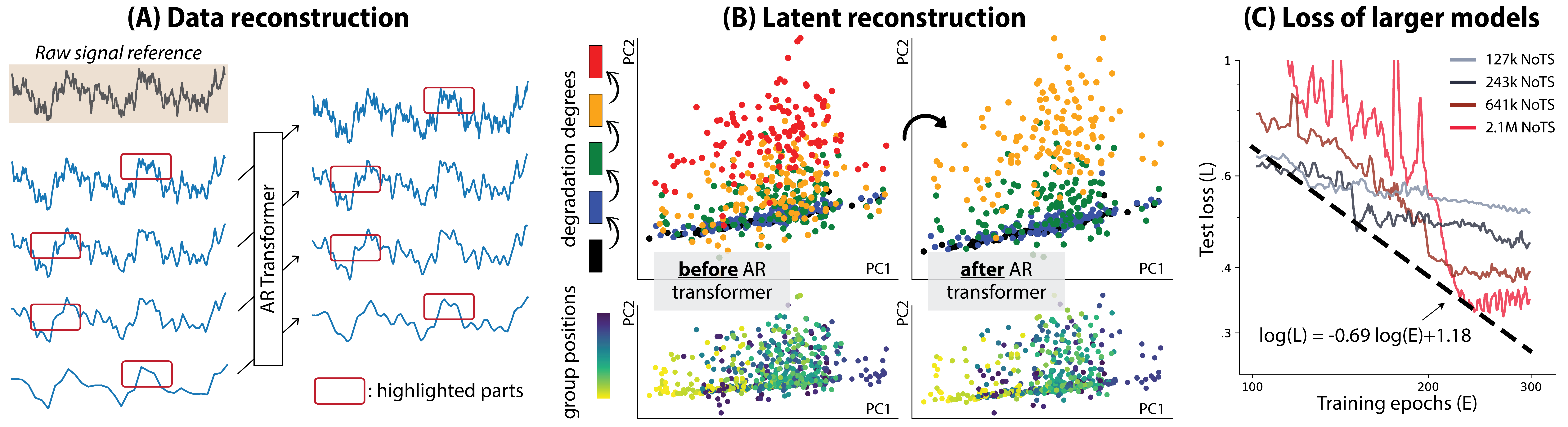}
  \vspace{-7mm}
  \caption{\footnotesize \textit{Visualizations of AR performance and loss}. (A) We visualize the autoregressive inference process of \ours~on the synthetic dataset. From bottom to top, the signal variance is gradually recovered through the prediction of the AR transformer. (B) The token space is visualized through principal component analysis, where tokens of the simplified signals gradually disperse to a larger region when colored in different degradation degrees. When colored with relative group positions, the distribution does not shift as much on the direction of another principal component. (C) A pilot study shows that training larger \ours~models leads to lower reconstruction loss on the test set, potentially following the power law behaviour of AR models.}
  \label{fig3:data_visualizations}
  \vspace{-4mm}
\end{figure}

\paragraph{Visualizing the next-function prediction process} We present the data and latent visualizations in Figure~\ref{fig3:data_visualizations}. In Figure~\ref{fig3:data_visualizations}(A), we show the original data sequence $\{\rmS_i\}_{i=1}^{K}$ and the reconstructed data sequence $\{\rmS_i^{\prime}\}_{i=2}^{K}$. Note that the original signal $\rmS_K$ was not passed into the transformer. We can see that the predicted sequence has information that is not presented in previous signals, showing the function prediction capacity of the transformer.
In Figure~\ref{fig3:data_visualizations}(B), we plot the token space before or after the AR transformer using the PCA reduction on $\{\rmR_i\}_{i=1}^{K}$ and $\{\rmR_i^{\prime}\}_{i=2}^{K}$, respectively. When coloring the tokens differently based on their degradation parameter $i$, we observe that: (1) In original token space $\{\rmR_i\}_{i=1}^{K}$, severely degraded signals generates more clustered tokens, and the tokens would gradually disperse as signals become more realistic; (2) The predicted tokens $\{\rmR_i^{\prime}\}_{i=2}^{K}$ would generate a token space with similar behaviour without seeing the original set of tokens $\rmR_K$. This behaviour demonstrates the autoregressive capacity of the transformer.

\begin{table}[t!]
\addtolength{\tabcolsep}{-0.2em}
  \begin{center}
    \caption{\footnotesize Comparisons between \ours~and other pre-training methods on real-world datasets. We categorize the results based on (a) if adaptors are used, and (b) if the weights of the pre-trained models are frozen. We compute an average error rate ($\downarrow$) to compare the final performance of different methods in each condition.
    }
    \label{tab:main}
    \vspace{-2mm}
\resizebox{\linewidth}{!}{%
\begin{tabular}{c?cc?cc?cccc?cccc?c} 
    & \multicolumn{2}{c}{} & \multicolumn{2}{c}{\textit{Classification ($\uparrow$)}} & \multicolumn{4}{c}{\textit{Anom. Det. ($\uparrow$)}} & \multicolumn{4}{c}{\textit{Imputation} ($\downarrow$)} & \textbf{Avg. ($\downarrow$)} \\
    \textbf{Methods} & (a) & (b) & UCR-9 & UEA-5 & SMD & MSL & SWaT & PSM & ETTm1 & ETTm2 & ETTh1 & ETTh2  & \textbf{error rate} \\
        \hline
\footnotesize{SimMTM} 
&\cmark&\cmark%
& 68.70 & 55.36 
& \cellcolor{cyan!10}\textbf{84.06} & 83.90 & 91.20 & 96.07 
& \cellcolor{cyan!10}\textbf{0.164} & 0.126 & 0.264 & 0.183 
& 19.43 \\
\footnotesize{bioFAME} 
&\cmark&\cmark%
& 62.63 & 60.32 
& 83.09 & \cellcolor{cyan!10}\textbf{84.28} & 91.21 & 95.94
& 0.203 & \cellcolor{cyan!10}\textbf{0.122} & \cellcolor{cyan!10}\textbf{0.258} & \cellcolor{cyan!10}\textbf{0.178}
& 19.87 \\
\footnotesize{Next-pred} 
&\cmark&\cmark%
& 65.95 & 58.30 
& 82.96 & 83.75 & 90.47 & \cellcolor{cyan!10}\textbf{96.54}
& 0.306 & 0.178 & 0.465 & 0.270
& 24.39 \\
\footnotesize{\textbf{\ourslw~(Ours)}} 
&\cmark&\cmark
& \cellcolor{cyan!10}\textbf{71.88} & \cellcolor{cyan!10}\textbf{62.78} 
& 83.63 & \cellcolor{cyan!10}\textbf{84.28} & \cellcolor{cyan!10}\textbf{93.26} & 96.27
& \cellcolor{cyan!10}\textbf{0.164} & 0.126 & 0.286 & 0.196
& \cellcolor{green!10}\textbf{18.51} \\
\hline
 \footnotesize{SimMTM} 
&\cmark&\xmark%
& 81.65 & 61.23 
& 83.48 & 84.11 & 91.35 & 96.36 
& 0.123 & \cellcolor{cyan!10}\textbf{0.107} & \cellcolor{cyan!10}\textbf{0.201} & 0.166
& 16.14 \\
\footnotesize{bioFAME} 
&\cmark&\xmark%
& 81.53 & 63.57 
& 83.59 & 83.98 & \cellcolor{cyan!10}\textbf{91.46} & \cellcolor{cyan!10}\textbf{96.88} 
& 0.129 & \cellcolor{cyan!10}\textbf{0.107} & 0.202 & 0.178
& 16.05 \\
\footnotesize{Next-pred} 
&\cmark&\xmark%
& 80.62 & 62.76 
& 83.00 & 84.09 & 91.00 & 96.87 
& 0.130 & 0.119 & 0.228 & 0.188
& 16.82 \\
\footnotesize{\textbf{\ourslw~(Ours)}} 
&\cmark&\xmark%
& \cellcolor{cyan!10}\textbf{88.08} & \cellcolor{cyan!10}\textbf{66.38}
& \cellcolor{cyan!10}\textbf{84.19} & \cellcolor{cyan!10}\textbf{84.15} & 91.26 & \cellcolor{cyan!10}\textbf{96.88} 
& \cellcolor{cyan!10}\textbf{0.122} & 0.116 & 0.218 & \cellcolor{cyan!10}\textbf{0.163} 
& \cellcolor{green!10}\textbf{15.10} \\
 \hline
\footnotesize{PatchTST} 
&\xmark&\xmark%
& 83.57 & 63.31
& 78.96 & 78.81 & 83.75 & 78.07
& 0.181 & 0.126 & 0.347 & 0.187
& 21.78 \\
\footnotesize{+\textbf{\ours~(Ours)}} 
&\xmark&\xmark%
& \cellcolor{cyan!10}\textbf{$\uparrow$1.71} & \cellcolor{cyan!10}\textbf{$\uparrow$1.65}
& \cellcolor{cyan!10}\textbf{$\uparrow$2.20} & \cellcolor{cyan!10}\textbf{$\uparrow$3.96} & \cellcolor{cyan!10}\textbf{$\uparrow$5.97} & \cellcolor{cyan!10}\textbf{$\uparrow$11.25} 
& \cellcolor{cyan!10}\textbf{$\downarrow$.003} & \cellcolor{cyan!10}\textbf{$\downarrow$.003} & \cellcolor{cyan!10}\textbf{$\downarrow$.064} & \cellcolor{cyan!10}\textbf{$\downarrow$.006}  
& \cellcolor{green!10}\textbf{18.33} \\
\footnotesize{iTransformer} 
&\xmark&\xmark%
& 82.67 & 67.62
& 85.18 & 83.04 & 91.88 & 97.07 
& 0.162 & 0.111 & 0.240 & 0.168 
& 16.07 \\
\footnotesize{+\textbf{\ours~(Ours)}} 
&\xmark&\xmark%
& \cellcolor{cyan!10}\textbf{$\uparrow$1.26} & \cellcolor{cyan!10}\textbf{$\uparrow$0.65}
& \cellcolor{cyan!10}\textbf{$\uparrow$0.17} & \cellcolor{cyan!10}\textbf{$\uparrow$0.11} & \cellcolor{cyan!10}\textbf{$\uparrow$0.05} & \cellcolor{cyan!10}\textbf{$\uparrow$0.01}
& $\uparrow$.005 & \cellcolor{cyan!10}\textbf{$\downarrow$.002} & \cellcolor{cyan!10}\textbf{$\downarrow$.013} & \cellcolor{cyan!10}\textbf{$\downarrow$.004}
& \cellcolor{green!10}\textbf{15.70} \\
    \end{tabular}}
    \vspace{-5mm}
  \end{center}
\end{table}


\vspace{-1mm}
\subsection{Real-world experiments: Context-aware generalization}
\vspace{-1mm}

\paragraph{Experimental setups}
To examine the performance of \ours~in real-world applications, we perform multi-task validation following the setups in \cite{wu2022timesnet}. Specifically, we perform the classification task on the UCR subset \citep{dau2019ucr} and UEA subset \citep{bagnall2018uea}; the imputation task on the ETDataset \citep{zhou2021informer}, and the anomaly detection task on MSL \citep{hundman2018detecting}, PSM \citep{abdulaal2021practical}, SWaT \citep{mathur2016swat}, and SMD \citep{su2019robust} datasets.
We follow \cite{wu2022timesnet} for standard data pre-processing and task deployment pipeline, except for the imputation task where we tested a more challenging variant of channel-wise imputation (see Appendix~\ref{app:imputation_details} and~\ref{app:additional_results} for details and original imputation results).

To validate that \ours~is a superior pre-training method, we perform two sets of experiments: 
First, we compare the performance of \ourslw~against the next-period AR transformer, a MAE \citep{dong2024simmtm}, and a frequency-aware MAE \citep{liu2023frequency} by pre-training them on synthetic datasets and deploying the prompt tuning pipeline for all pre-trained base models.
Second, we append the pre-training pipeline \ours~on top of existing architectures PatchTST \citep{nie2022time} and iTransformer \citep{liu2023itransformer}, and compute the performance benefits from adding \ours.

\paragraph{Experimental results} As shown in Table~\ref{tab:main}, with or without parameters frozen, \ourslw~significantly outperforms all other pre-training methods. Specifically, given the same pre-training pipeline and architecture, \ourslw~outperforms other method across all tasks by up to $6\%$ average. Interestingly, we note that \ourslw~show comparable performance on imputation tasks, where MAE-like architectures are trained to perform the task. Additionally, when attaching \ours~on existing architectures PatchTST \citep{nie2022time} and iTransformer \citep{liu2023itransformer}, \ours~improves their performance without specific backbone or adaptors, showing the versatility of the pre-training method.

Interestingly, we should like to emphasize on the context-aware generalization ability of \ours. With the architecture frozen (first 4 rows of Table~\ref{tab:main}), we only train \textless $1\%$ of the parameters, yet it performs $82\%$ performance, potentially demonstrating the context-aware generalization.


\subsection{Ablation experiments and model analysis}



\paragraph{Ablation of effective components in \ours} In Table~\ref{tab:ablations}, we perform ablations of \ours~by isolating the effective components of \ourslw~in the feature regression task (the $\mathcal{H}$-index). Specifically, we train three variants of \ourslw~by (1) removing the latent consistency term in training loss, (2) removing the autoregressive masking within transformer, creating a transformer that merely bridges tokens of the augmented samples, (3) removing the connections among constructed augmentations, and using degradation operator only as augmentations. As expected, removing the latent consistency term would cause distributional shift as the model never sees raw data, and would result in severe performance degradation. Interestingly, training a transformer that connects augmented samples can also provide improved performance, as observed in other works \citep{hou2022batchformer,liu2024latentdr}.

\begin{wraptable}{r}{5.3cm}
\vspace{-7.5mm}
\caption{\footnotesize Ablation experiments. Columns denote: If the model sees original signal (orig.), if the model uses autoregressive modeling (AR), if the signal variants are connected (conn.), and if a Gaussian-based degradation operator is used ($d_k^{(\mathcal{N})}$).}
\label{wrap-tab:1}
\begin{center}
\addtolength{\tabcolsep}{-0.4em}
\resizebox{5cm}{!}{%
\begin{tabular}{ccccc?c}
& orig. & AR & conn. & $d_k^{(\mathcal{N})}$ & error ($\downarrow$) \\ \hline
(1) & \xmark & \cmark & \cmark & \xmark & 1.75 \\
(2) & \cmark & \xmark & \cmark & \xmark & 1.48 \\
(3) & \cmark & \xmark & \xmark & \xmark & 1.82 \\
(4) & \cmark & \cmark & \cmark & \cmark & 1.69 \\
\hline
\ours & \cellcolor{cyan!10}\cmark & \cellcolor{cyan!10}\cmark & \cellcolor{cyan!10}\cmark & \cellcolor{cyan!10}\xmark & \cellcolor{cyan!10}1.27 \\
\end{tabular}}
\label{tab:ablations}
\vspace{-3mm}
\end{center}
\end{wraptable} 

\vspace{-2mm}
\paragraph{Connection to diffusion models} One might relate our work with diffusion models \citep{ho2020denoising} by using a stochastic additive Gaussian noise of varying degrees as a degradation operator. We attempted this as model variant (4) in Table~\ref{tab:ablations}, yet the performance is inferior to the convolution-based degradation operators. One hypothesis is that time series data is inherently noisy, and adding Gaussian noise instead of performing smoothing or filtering can be less effective, as observed in audio signals \citep{dieleman2024spectral}. Building connections between \ours~and cold diffusion models with deterministic degradation operators \citep{bansal2024cold}, or more recent diffusion models \citep{chen2023generative} can be an exciting future research direction.


\paragraph{Scalability analysis} While this work aims only to provide an initial experimental exploration of the proposed pre-training methodology \ours, we attempted a pilot study to increase the size of \ourslw~to demonstrate its potential given more parameters. We trained four models with 127k, 243k (used in all previous experiments), 641k, 2.1M parameters to observe their performance. As shown in Figure \ref{fig3:data_visualizations}(C), when fixing the amount of training data, training the models to convergence with increased parameters leads to increased performance, potentially following a power law curve of AR frameworks in language and computer vision \citep{kaplan2020scaling,el2024scalable}.



\vspace{-2mm}
\section{Conclusion}
\vspace{-2mm}
In this paper, we propose a novel autoregressive pre-training method \ours~for time series. Our work aims to provide an alternative view of time series by considering them as functions of time instead of concatenations of time periods. This novel perspective allows us to construct degradation operators, which build an alternative sequence as inputs to the transformer.
The transformer is pre-trained with an autoregressive loss to encourage the learning of cross-function relationship, building a model that can recover signal variability from their simplified variants.
We validated the performance of \ours~with experimental results on 2 synthetic and 22 real-world datasets, demonstrating its superiority among existing pre-training methods across multiple tasks, showing a viable alternative for developing foundation models for time series analysis in the future.

\vspace{-2mm}
\paragraph{Limitations} Future works may extend the existing results through: (1) Expanding our initial experimental efforts to larger models, larger-scale datasets, and more challenging tasks. (2) Building in-depth theoretical connection to diffusion-based models, connecting \ours~with recent works \citep{dieleman2024spectral} from the audio and computer vision domain. (3) Understanding how \ours~performs 
in stochastic events as detailed in \citep{kidger2020neural} based on rough path theory. 


\subsubsection*{Acknowledgments}

We gratefully acknowledge Apple for supporting this project: This work was supported by a grant from Apple, Inc. Any views, opinions, findings, and conclusions or recommendations expressed in this material are those of the authors and should not be interpreted as reflecting the views, policies or position, either expressed or implied, of Apple Inc.
We also acknowledge the NSF CAREER Award (IIS-2146072) and the CIFAR Azrieli Global Scholars Program, which supported Dr. Dyer’s contributions to this work.
We are grateful to all data contributors, including all the people who have contributed to the UCR and the UEA time series archives; the iTrust Centre for Research in Cyber Security at the Singapore University of Technology and Design; and all others. We also thank Keyu Chen and Tianrong Chen for their inspiring discussions.

\bibliography{biofeature}
\bibliographystyle{iclr2025_conference}

\newpage
\appendix
\section*{Appendix}

\section{Analytical results}
\label{app:theorysec}

\subsection{Preliminaries}

\paragraph{Transformers} We consider the standard transformer architecture as defined in \citep{luo2022your}.
The transformer network is the stack of transformer blocks, each of them consists of a self-attention layer $\operatorname{Attn}(\cdot)$ and a feed forward layer $\operatorname{FF}(\cdot)$. 
Given an input $\rmX \in \mathbb{R}^{d \times T}$,
they are written as:
\begin{equation}
\begin{aligned}
\label{app:attention_basics}
\operatorname{Attn}(\boldsymbol{X}) & =\boldsymbol{X}+\sum_{i=1}^h \boldsymbol{W}_O^i \boldsymbol{W}_V^i \boldsymbol{X} \cdot \sigma\left[\left(\boldsymbol{W}_K^i \boldsymbol{X}\right)^T \boldsymbol{W}_Q^i \boldsymbol{X}\right] \\
\mathrm{FF}(\boldsymbol{X}) & =\operatorname{Attn}(\boldsymbol{X})+\boldsymbol{W}_2 \cdot \operatorname{ReLU}\left(\boldsymbol{W}_1 \cdot \operatorname{Attn}(\boldsymbol{X})\right)
\end{aligned}
\end{equation}
where $\boldsymbol{W}_O^i \in \mathbb{R}^{d \times m}$, $\boldsymbol{W}_V^i, \boldsymbol{W}_K^i, \boldsymbol{W}_Q^i \in \mathbb{R}^{m \times d}$, $\boldsymbol{W}_2 \in \mathbb{R}^{d \times r}$, and $\boldsymbol{W}_1 \in \mathbb{R}^{r \times d}$.

We denote $t^{h, m, r}: \mathbb{R}^{d \times T} \rightarrow \mathbb{R}^{d \times T}$ as a transformer block with an attention layer with $h$ heads of size $m$, and a feed-forward layer with $r$ hidden nodes. Thus, the transformer can be written as:
\begin{equation}
\mathcal{T}^{h, m, r}:=\left\{f: \mathbb{R}^{d \times T} \rightarrow \mathbb{R}^{d \times T} \mid f \text { is a composition of transformer blocks } t^{h, m, r} \right\} .
\end{equation}
Similarly, transformer with absolute positional embedding is:
\begin{equation}
\mathcal{T}_{\mathrm{P}}^{h, m, r}:=\left\{f_{\mathrm{P}}(\boldsymbol{X})=f(\boldsymbol{X}+\boldsymbol{E}) \mid f \in \mathcal{T}^{h, m, r} \text { and } \boldsymbol{E} \in \mathbb{R}^{d \times T}\right\}
\end{equation}

\paragraph{Universal Approximator (UA)}
The universal approximation framework considers the feasibility or existence of a neural network that can approximate different types of functions with arbitrarily small error. 
Consider a transformer network $f_1$ and an arbitrary function $f_2$, where $f_1, f_2: \mathbb{R}^{n \times T} \rightarrow \mathbb{R}^{n \times T}$ are both sequence-to-sequence functions. We define a distance between $f_1$ and $f_2$ as:
\begin{equation}
\mathrm{d}_p\left(f_1, f_2\right):=\left(\int\left\|f_1(\boldsymbol{X})-f_2(\boldsymbol{X})\right\|_p^p d \boldsymbol{X}\right)^{1 / p} 
\end{equation}
being a UA means that for any given $f_2 \in \mathcal{F}_{}$, let $1 \leq p<\infty$ and $\epsilon>0$, there exists a network $f_1$, such that $\mathrm{d}_p(f_1, f_2) \leq \epsilon$.
Several prior works have explored the concept of universal approximators (UAs) and whether transformers qualify as UAs. Below, we outline the key results from the literature that will be referenced in this paper:

\textbf{Theorem 2} (informal, see \cite{yun2019transformers}).
Given $1 \leq p<\infty$ and $\epsilon>0$, for any function $f \in \mathcal{F}_{\mathrm{PE}}$, where $\mathcal{F}_{\mathrm{PE}}$ consists of all continuous permutation equivariant functions with compact support, there exists a Transformer network $f \in \mathcal{T}^{2,1,4}$ where $\mathrm{d}_p(f, g) \leq \epsilon$.

\textbf{Theorem 3} (informal, see \cite{yun2019transformers}).
Given $1 \leq p<\infty$ and $\epsilon>0$, for any function $f \in \mathcal{F}_{\mathrm{CD}}$, where $\mathcal{F}_{\mathrm{CD}}$ consists of all
continuous functions with compact support, there exists a Transformer network $f \in \mathcal{T}_{\mathrm{P}}^{2,1,4}$ where $\mathrm{d}_p(f, g) \leq \epsilon$.

Theorem 2 discussed that transformers without positional embeddings are UAs for all continuous permutation equivariant functions; and Theorem 3 discussed that transformers with absolute positional embeddings (APE) are UAs for all continuous functions with compact support. Note that the latter results may be overruled by modifying the transformer architectures as follows:


\textbf{Theorem 4} (informal, see \cite{luo2022your}). Consider relative positional encoding (RPE) that modifies the attention as $\operatorname{Attn}(\boldsymbol{X}) = \boldsymbol{X}+\sum_{i=1}^h \boldsymbol{W}_O^i \boldsymbol{W}_V^i \boldsymbol{X} \cdot \operatorname{softmax}\left[\left(\boldsymbol{W}_K^i \boldsymbol{X}\right)^T \boldsymbol{W}_Q^i \boldsymbol{X} + \rmE \right]$, where $\rmE \in \mathbb{R}^{T \times T}$ encodes the relative position within attention maps. Given $T>2$, there always exists a continuous function $f_M: \mathcal{D} \to \mathbb{R}^{d \times T}$, such that $\sup _{\boldsymbol{X} \in \mathcal{D}}\left\|f_{\operatorname{RPE}}(\boldsymbol{X})-f_{M}(\boldsymbol{X})\right\|_2^2 \geq M$ holds for any modified RPE-based transformer.

While UA framework is typically used for understanding the approximation problem towards continuous functions, more recently, it is used to understand the approximation problem towards discontinuous functions. Specifically, in this work, we reference the results in \cite{ismailov2023three} that shows any discontinuous function may be implemented by a three-layer Kolmogorov type neural network:

\textbf{Theorem 5} (informal, see \cite{ismailov2023three}). Given $d \geq 2$ and any function $f : \mathbb{I}^{d} \to \mathbb{R}$, where $\mathbb{I}$ is a closed unit internal $[0, 1]$, then function $f$ can be implemented exactly by a three-layer Kolmogorov neural network with $d$, $2d+1$, and $1$ processing units in the first, second, and final layer, respectively. 


As stated in \cite{ismayilova2023kolmogorov,ismailov2023three}, the expressiveness of simple neural networks can be extended by constructing more diverse activation functions, which helps us understand the approximation ability towards discontinuous functions which are more prevalent in real-world complex systems. Note that other works also discuss the approximation problems towards functions that may be discontinuous \citep{kidger2020universal,pinkus1999approximation}.


\subsection{Additional proof of theorem 1}
\label{app:lemma1}

We only study the convergence property of self attention layers as the convergence property of feed forward networks has been extensively studied in previous works.
To prove convergence, we build an input sequence $\rmX  + \Delta_n$, where $\Delta_n$ is defined as a bounded perturbation matrix $\Delta \in \mathcal{D}$ that is uniformly scaled by a positive value $n$. Given a self attention layer $\operatorname{Attn}(\rmX)$, we show the following:

\textbf{Lemma 1.} Given $n \geq N$, $\rmX \in \mathcal{D}$, $\Delta_n = \Delta/n$, there exists an $\epsilon$ such that:
\begin{equation}
\label{app:eq10}
    \sup_{\Delta \in \mathcal{D}, \|\Delta\|_1 \leq 1} \| \operatorname{Attn} (\rmX) - \operatorname{Attn}(\rmX  + \Delta_n)\|_2 < \epsilon.
\end{equation}
holds for any self attention layer parameterized by $\boldsymbol{W}_O^i \in \mathbb{R}^{d \times m}$, $\boldsymbol{W}_V^i, \boldsymbol{W}_K^i, \boldsymbol{W}_Q^i \in \mathbb{R}^{m \times d}$.

\textbf{Proof.} First, we re-write the activation component $\sigma\left[\left(\boldsymbol{W}_K^i \boldsymbol{X}\right)^\top \boldsymbol{W}_Q^i \boldsymbol{X}\right]$ in Eq.~\ref{app:attention_basics} as column-wise softmax operation $\operatorname{softmax}(\rmX^\top \rmW \rvx_j)$, where $\rvx_j$ is a random column of $\rmX$. We have:
\begin{equation}
\begin{aligned}
\| & \operatorname{softmax}(\rmX^\top \rmW \rvx_j) - \operatorname{softmax}((\rmX+\Delta_n)^\top \rmW (\rvx_j + \delta_{j,n}))\|_2 \\
& \leq \| \operatorname{softmax}(\rmX^\top \rmW \rvx_j) - \operatorname{softmax}((\rmX+\Delta_n)^\top \rmW (\rvx_j + \delta_{j,n}))\|_1 \\
& \leq 2 \|\rmX^\top \rmW \rvx_j - (\rmX+\Delta_n)^\top \rmW (\rvx_j + \delta_{j,n})\|_{\infty} \text{ (Corollary A.7 in \cite{edelman2022inductive})} \\
& = 2 \max_{i} (\frac{1}{n} (\rvx_i^\top \rmW \delta_{j} + \delta_i^\top \rmW \rvx_j) + \frac{1}{n^2} \delta_i^\top \rmW \delta_j) \\
& \leq 2 \max_{i} (\frac{1}{n} (\rvx_i^\top \rmW \mathds{1} + \mathds{1}^\top \rmW \rvx_j) + \frac{1}{n^2} \mathds{1}^\top \rmW \mathds{1}) = \epsilon_h
\end{aligned}
\end{equation}
which shows that the attention map converges given deterministic $\rmX$ and $\rmW = (\rmW_K^i)^\top \rmW_Q^i$. Thus, Eq.~\ref{app:eq10} holds by considering the self attention operator $\operatorname{Attn} (\rmX)$ as a convex combination of attention heads given deterministic $\rmX$, $\boldsymbol{W}_O^i \in \mathbb{R}^{d \times m}$, $\boldsymbol{W}_V^i \in \mathbb{R}^{m \times d}$.

Thus, given $\boldsymbol{X} \rightarrow \mathbf{0}$, the transformer network $f_P(\rmX)$ converges to a deterministic matrix $\rmB$ that is dependent on the network parameters and the positional embedding $\rmE$.
Note that, under the context of network optimization, a more generalized version of the convergence property of transformers has been proved in other previous works \citep{wu2024convergence,gaoglobal}. 




\subsection{Additional proof of proposition 1}
\label{app:proposition}

Based on the results in Theorem 3, it is known that as long as the constructed sequence $\rmX = [\mathcal{E}(\rmS_1), \mathcal{E}(\rmS_2), ..., \mathcal{E}(\rmS_T)]$ forms a continuous sequence-to-sequence function between input $\rmX$ and the target $\{A[g(t_i)]\}_{i=1}^{T}$, it is guaranteed that there exists a transformer network $f_P \in \mathcal{T}_P^{2, 1, 4}$ that can approximate the constructed sequence-to-sequence function. 
Thus, we show how the presented two conditions are sufficient to meet the above requirement:
\begin{itemize}[noitemsep,leftmargin=*,topsep=0pt]
    \item When there exists a continuous mapping between a fixed element $p$ of $\rmS_i$ and the $i$-th element of the target output $A[g(t_i)]$, one can construct a simple linear encoder $\mathcal{E}(\rmS) = \rmS \rvv$, where $\rvv[i] = 0$ when $i \neq p$ and $\rvv[p] = 1$, that creates the continuous sequence-to-sequence function.
    \item Based on the results in Theorem 5,  if there exists an expressive tokenizer $ \mathcal{E}$ (that may be a discontinuous function) that preprocess $\rmS_i$ to create a continuous mapping between $\mathcal{E}(\rmS_i)$ to the target, the existence of the transformer is guaranteed for a continuous sequence-to-sequence function.
\end{itemize}

\section{Experimental details}

\subsection{Synthetic experiments}
\label{app:synthetic_exp}

\subsubsection{Synthetic datasets details}
\paragraph{Fractional Brownian motion (fBm)} 
Given a Hurst index $\mathcal{H}$ and two time steps $i$ and $j$, a fBm process is a continuous-time Gaussian process $B_\mathcal{H}(t)$ with the following covariance structure:
\begin{equation}
E\left[B_\mathcal{H}(i) B_\mathcal{H}(j)\right]=\frac{1}{2}\left(|i|^{2 \mathcal{H}}+|j|^{2 \mathcal{H}}-|i-j|^{2 \mathcal{H}}\right)
\end{equation}
Define function $\gamma(i, \mathcal{H}) = 0.5(|i-1|^{2\mathcal{H}}+|i+1|^{2\mathcal{H}}-2|i|^{2\mathcal{H}})$, a fBm process can be simulated through the Cholesky decomposition method detailed as follows:


\begin{algorithm}[H]
\caption{Simulation of fBm processes using the Cholesky's method}
\label{alg:fbm_generation}
\begin{algorithmic}
\State \textbf{Inputs:} $N$ as the length of sequence (time steps), $\mathcal{H} \in (0, 1)$ as the Hurst index
\State \textbf{Initialize:} $\rmL \in \mathbb{R}^{N \times N}$, $\rmV \in \mathbb{R}^N$ with each entry randomly sampled from $\mathcal{N}(0, 1)$
\State \textbf{Define:} $\rmX \in \mathbb{R}^N$ as the output vector

Initial conditions for $\rmL$: $\rmL[0,0] = 1$, $\rmL[1,0] = 2^{2\mathcal{H}-1} - 1$, $\rmL[1,1] = (1 - \rmL[1,0]^2)^{1/2}$,

Initial conditions for $\rmX$: $\rmX[0] = \rmV[0]$, $\rmX[1] = \rmL[1,0] \rm\rmV[0] + \rmL[1,1] \rmV[1]$
\For{each time step $i$ from $2$ till $N-1$}
    $\rmL[i,0] = \gamma(i, \mathcal{H})$
    \For{each time step $j$ from $1$ to $i-1$}
        \[
        \rmL[i,j] = \frac{1}{\rmL[j,j]} \left( \gamma(i-j, \mathcal{H}) - \sum_{k=0}^{j-1} \rmL[i,k] \cdot \rmL[j,k] \right)
        \]
    \EndFor
    
Update $\rmL[i,i] = (1 - \sum_{k=0}^{i-1}(\rmL[i,k]^2))^{1/2}$, 
$X[i] = \sum_{k=0}^{i} \rmL[i,k] \rmV[k]$
\EndFor

\For{each time step $i$ from $N-1$ till $0$}
 $\rmX[i] = (\sum_{k=0}^{i} \rmX[k]) \times N^{-\mathcal{H}}$
\EndFor

\State \textbf{Output:} A simulated fBm process $\rmX$

\end{algorithmic}
\end{algorithm}

\vspace{-3mm}
\paragraph{Autocorrelated sinusoids}

The autocorrelated sinusoids dataset is generated with AR processes in the frequency space. Given an integer $k$, a randomly initialized set of weights $\{\phi_i\}_{i=1}^k$, an AR($k$) process defines the sequence of frequency values as follows:
\begin{equation}
    f_t = \sum_{i=1}^{k} \phi_i f_{t-i}
\end{equation}
The AR process ensures that the frequency components in the synthetic dataset are correlated, creating an autoregressive frequency structure that \ours~can effectively learn from.

\subsubsection{Feature regression task details}

We detail the feature extraction methods as follows. We use them as the ground truth for the feature regression task.
Define the indicator function $\mathbf{1}_A(x)$, where $\mathbf{1}_A(x) = 1$ if $x \in A$ and $\mathbf{1}_A(x) = 0$ otherwise. Given a single-channel signal $\rvs \in \mathbb{R}^{T}$ with $v_i$ as the value on $i$-th timestamp, all features are extracted on each channel of the signal as follows:

\paragraph{Slope Sign Change (SSC)} SSC measures directional slope changes in a signal, indicating the intensity of fluctuations. Given a threshold value $\delta$ as hyperparameter, a period of time series sequence $\rvs$, we extract SSC value with the following equation:
\[\operatorname{SSC}(\rvs) = \sum_{i=2}^{T-1} \mathbf{1}_{(v_i - v_{i-1})(v_{i} - v_{i+1}) < 0} \left( v_i \right) \cdot \mathbf{1}_{\max \left( |v_{i} - v_{i+1}|, |v_{i} - v_{i-1}| \right) \geq \delta} \left( v_i \right) \] 
In practice, we extract the SSC values on top of segmented signals with a length of $32$.

\paragraph{Willison Amplitude (WAMP)} WAMP is a similar feature that focuses on counting significant amplitude changes between consecutive steps. Given a threshold value $\delta$ as hyperparameter, a period of time series sequence $\rvs$, WAMP is computed through $\operatorname{WAMP}(\rvs) = \sum_{i=1}^{T-1} \mathbf{1}_{|v_{i+1} - v_{i}| \geq \delta} \left( v_i \right)$. In practice, we also extract the WAMP values on top of segmented signals with a length of $32$, creating a 32-dimensional feature for each studied synthetic data sample.

\paragraph{Band power (b. power)} The band power quantifies the energy within a specific selected range of frequencies. It is computed by first performing the Fourier transform of $\rvs$, creating a frequency representation $\rvs(f)$. The band power within frequency range $[f_1, f_2]$ is later extracted as $\operatorname{BP_{(f_1, f_2)}}(\rvs) = \int_{f_1}^{f_2} |\rvs(f)|^2 \, df$.
In this work, we consider 3 unique frequency range $\{[5, 10], [15, 30], [30, 80]\}$ as hyperparameters to extract a 96-dimensional feature for each studied synthetic data sample.

\subsection{Real-world experiments}

\subsubsection{Dataset information}
\paragraph{Classification}

We selected 9 univariate datasets from the UCR archive \citep{dau2019ucr}, filtering out all datasets with less than $140$ series length or less than $350$ training samples. The dataset selection is performed to ensure each dataset has both sufficient samples and dynamics.
The detailed information about the selected datasets is provided in Table~\ref{table:selected_classification_ucr}.

\begin{table}[htbp]
    \centering
    \footnotesize
    \begin{tabular}{l c c c c c}
    \toprule
    \textbf{Dataset} & \textbf{Train} & \textbf{Test} & \textbf{Series Length} & \textbf{Classes} \\
    \midrule
    FordA                  & 3601  & 1320  & 500  & 2 \\
    \cmidrule(lr){1-5}
    FordB                  & 3636  & 810   & 500  & 2 \\
    \cmidrule(lr){1-5}
    ScreenType             & 375   & 375   & 720  & 3 \\
    \cmidrule(lr){1-5}
    ECG5000                & 500   & 4500  & 140  & 5 \\
    \cmidrule(lr){1-5}
    Wafer                  & 1000  & 6164  & 152  & 2 \\
    \cmidrule(lr){1-5}
    StarLightCurves        & 1000  & 8236  & 1024 & 3 \\
    \cmidrule(lr){1-5}
    UWaveGestureLibraryAll & 896   & 3582  & 945  & 8 \\
    \cmidrule(lr){1-5}
    HandOutlines           & 1000  & 370   & 2709 & 2 \\
    \cmidrule(lr){1-5}
    EthanolLevel           & 504   & 500   & 1751 & 4 \\
    \bottomrule
    \end{tabular}
    \caption{Detailed information about the selected datasets from the UCR archive.}
    \label{table:selected_classification_ucr}
\end{table}

We also selected 5 multivariate datasets from the UEA archive \citep{bagnall2018uea}, excluding those with a series length below $100$ and the training sample size below $200$. The detailed information about the selected datasets is provided in Table~\ref{table:selected_classification_uea}.

\begin{table}[htbp!]
    \centering
    \footnotesize
    \begin{tabular}{l c c c c c c}
        \toprule
        \textbf{Dataset} & \textbf{Channel} & \textbf{Train} & \textbf{Test} & \textbf{Series Length} & \textbf{Classes} \\
        \midrule
        EthanolConcentration   & 3   & 261   & 263   & 1751 & 4 \\
        \cmidrule(lr){1-6}
        Heartbeat              & 61  & 204   & 205   & 405  & 2 \\
        \cmidrule(lr){1-6}
        PEMS-SF                & 963 & 267   & 173   & 144  & 7 \\
        \cmidrule(lr){1-6}
        SelfRegulationSCP1     & 6   & 268   & 293   & 896  & 2 \\
        \cmidrule(lr){1-6}
        SelfRegulationSCP2     & 7   & 200   & 180   & 1152 & 2 \\
        \bottomrule
    \end{tabular}
    \caption{Detailed information about the selected datasets from the UEA archive.}
    \label{table:selected_classification_uea}
\end{table}

\paragraph{Imputation}
For the imputation tasks, we use the ETDataset \citep{zhou2021informer}, where ETTm1 and ETTm2 are sampled at minute intervals, and ETTh1 and ETTh2 are sampled at hourly intervals. The detailed information about the selected datasets is provided in Table~\ref{table:ett_datasets}.

\begin{table}[htbp]
    \centering
    \footnotesize
    \begin{tabular}{l c c c c c}
        \toprule
        \textbf{Dataset} & \textbf{Channel} & \textbf{Series Length} & \textbf{Train} & \textbf{Validation} & \textbf{Test} \\
        \midrule
        ETTm1, ETTm2  & 7  & 96 & 34465  & 11521  & 11521 \\
        \cmidrule(lr){1-6}
        ETTh1, ETTh2  & 7  & 96 & 8545   & 2881   & 2881 \\
        \bottomrule
    \end{tabular}
    \caption{ETDataset for imputation tasks.}
    \label{table:ett_datasets}
\end{table}

\paragraph{Anomaly detection} The detailed information about the selected datasets is provided in Table~\ref{table:anomaly_detection_datasets}.

\begin{table}[htbp!]
    \centering
    \footnotesize
    \begin{tabular}{l c c c c c}
        \toprule
        \textbf{Dataset} & \textbf{Channel} & \textbf{Series Length} & \textbf{Train} & \textbf{Validation} & \textbf{Test} \\
        \midrule
        SMD   & 38  & 100 & 566724  & 141681  & 708420 \\
        \cmidrule(lr){1-6}
        MSL   & 55  & 100 & 44653   & 11664   & 73729 \\
        \cmidrule(lr){1-6}
        SWaT  & 51  & 100 & 396000  & 99000   & 449919 \\
        \cmidrule(lr){1-6}
        PSM   & 25  & 100 & 105984  & 26497   & 87841 \\
        \bottomrule
    \end{tabular}
    \caption{Detailed information about the selected datasets for the anomaly detection tasks.}
    \label{table:anomaly_detection_datasets}
\end{table}

\vspace{-4mm}
\subsubsection{Model training and architecture details}
\label{app:imputation_details}

\paragraph{Training details} For pre-training on synthetic datasets, we use a learning rate of $0.05$, a MultiStepLR scheduler with a multiplicative factor $\gamma=0.3$, and two milestones on epoch $30$ and $150$. We perform all pre-training for a total of $300$ epochs on both of the synthetic datasets, where we set batch size as $1024$ for the reconstruction task.

For pre-training on real-world datasets, we use a learning rate of $0.005$. We perform all pre-training for either a total of $300$ epochs, or a total of $6000$ steps, whichever finishes the first. We set batch size as $32$ for imputation and anomaly detection tasks, and batch size as $64$ for classification tasks.

\begin{table}[b!]
\addtolength{\tabcolsep}{-0.2em}
\footnotesize
\vspace{-4mm}
  \begin{center}
    \caption{Complete classification results on the UCR datasets.}
    \label{tab:app_clf_ucr}
\resizebox{\linewidth}{!}{%
\begin{tabular}{c?cccc?cccc} 
    & \multicolumn{4}{c?}{\textbf{Parameter efficient tuning}} & \multicolumn{4}{c}{\textbf{Full-scale fine-tuning}} \\
   \textbf{Dataset}  & \textbf{NoTS-lw}  & {Next-pred}  & {bioFAME} & {SimMTM} & \textbf{NoTS-lw} & {Next-pred} & {bioFAME} & {SimMTM} \\
    \hline
    HandOutlines & 71.62 & 64.32 & 64.05 & 88.92 & 93.51 & 72.16 & 91.62 & 89.73 \\
    EthanolLevel & 28.60 & 26.60 & 25.20 & 29.00 & 91.40 & 48.60 & 41.20 & 38.00 \\
    StarLightCurves & 91.66 & 87.34 & 85.15 & 88.59 & 97.21 & 97.49 & 97.56 & 97.39 \\
    UWave-GL-All & 67.39 & 56.28 & 37.38 & 76.35 & 96.57 & 96.90 & 87.83 & 94.72 \\
    FordA & 81.52 & 77.27 & 71.74 & 51.59 & 94.02 & 94.09 & 93.49 & 94.55 \\
    FordB & 68.27 & 63.83 & 57.41 & 64.57 & 83.70 & 86.17 & 85.19 & 83.58 \\
    Wafer & 98.78 & 89.21 & 89.21 & 89.23 & 99.81 & 99.87 & 99.64 & 99.85 \\
    ECG5000 & 91.31 & 89.00 & 92.96 & 87.38 & 94.13 & 88.47 & 94.29 & 93.82 \\
    ScreenType & 47.73 & 39.73 & 40.53 & 42.67 & 42.40 & 41.87 & 42.93 & 43.20 \\
    \hline
    \textbf{Average} & 71.88 & 65.95 & 62.63 & 68.70 & 88.08 & 80.62 & 81.53 & 81.65 \\
\end{tabular}}

\begin{tabular}{c?cccc} 
    \multicolumn{5}{c}{} \\
    & \multicolumn{4}{c}{\textbf{Attaching NoTS to existing architectures}} \\
   \textbf{Dataset}  & PatchTST  & PatchTST + \textbf{NoTS}  & {iTransformer} & iTransformer + \textbf{NoTS} \\
    \hline
    HandOutlines & 91.89 & 93.51 & 92.16 & 92.16 \\
    EthanolLevel & 57.80 & 68.00 & 86.20 & 86.40 \\
    StarLightCurves & 97.46 & 97.56 & 93.94 & 93.52 \\
    UWave-GL-All & 96.04 & 96.45 & 89.89 & 91.76 \\
    FordA & 93.71 & 93.56 & 77.05 & 83.11 \\
    FordB & 78.64 & 80.49 & 68.52 & 69.14 \\
    Wafer & 99.59 & 99.63 & 99.72 & 99.77 \\
    ECG5000 & 94.09 & 94.33 & 94.42 & 94.49 \\
    ScreenType & 42.93 & 44.00 & 42.13 & 45.07 \\
    \hline
    \textbf{Average} & 83.57 & 85.28 & 82.67 & 83.94 \\
\end{tabular}
  \end{center}
\end{table}

We apply the same set of hyperparameters for both parameter efficient fine-tuning and full-scale fine-tuning, where we perform hyperparameter selection on learning rate $\{0.005, 0.001, 0.05\}$ and batch size $\{32, 64, 128\}$. We perform the fine-tuning for 300 epochs on imputation, anomaly detection, and feature regression tasks, and perform the fine-tuning for $4000$ steps on classification tasks.

The settings are applied consistently across all models to ensure a fair comparison. All models are optimized with an Adam optimizer with $\beta_1 = 0.9$, $\beta_2 = 0.99$, and a weight decay of $1 \times 10^{-5}$.

\paragraph{Model architectures} 
For all pre-training methods including \ourslw, we use a same channel-independent 1D-ResNet encoder for fair comparison. The encoder has $3$ ResNet layers of channel size $\{16, 32, 64\}$, each has $2$ ResNet blocks. The first convolutional layer has a kernel size of $7$, and the rest layers have a kernel size of $3$.
We append an additional convolutional layer after the ResNet blocks to alter the dimensionality $d$ of the token embeddings, where model variant $d=32$ is used for all experiments, and $d=16, 64, 128$ is trained for the scalability pilot study. 
We use a $3$-layer $4$-head transformer with a token dimension of $d$, and $4 \times$ size in the feed forward layer. 
The decoder is built to be symmetric to the encoder architecture.

\begin{table}[h!]
\addtolength{\tabcolsep}{-0.4em}
\footnotesize
\vspace{-3mm}
  \begin{center}
    \caption{Complete classification results on the UEA datasets.}
    \label{tab:app_clf_uea}
\resizebox{\linewidth}{!}{%
\begin{tabular}{c?cccc?cccc} 
    & \multicolumn{4}{c?}{\textbf{Parameter efficient tuning}} & \multicolumn{4}{c}{\textbf{Full-scale fine-tuning}}  \\
   \textbf{Dataset}  & \textbf{NoTS-lw}  & {Next-pred}  & {bioFAME} & {SimMTM} & \textbf{NoTS-lw} & {Next-pred} & {bioFAME} & {SimMTM} \\
    \hline
  EthanolConcentration & 28.14 & 25.48 & 28.14 & 25.48 & 30.04 & 29.28 & 27.76 & 28.90 \\
    Heartbeat & 73.66 & 73.17 & 72.68 & 72.20 & 74.63 & 73.66 & 73.17 & 73.17 \\
    PEMS-SF & 75.72 & 58.38 & 77.46 & 64.16 & 80.35 & 67.63 & 75.15 & 72.25 \\
    SelfRegulationSCP1 & 79.18 & 77.82 & 69.97 & 59.39 & 89.08 & 86.01 & 85.67 & 74.06 \\
    SelfRegulationSCP2 & 57.22 & 56.67 & 53.33 & 55.56 & 57.78 & 57.22 & 56.11 & 57.78 \\
        \hline
    \textbf{Average} & 62.78 & 65.95 & 62.63 & 68.70 & 88.08 & 80.62 & 81.53 & 81.65 \\
\end{tabular}}
\begin{tabular}{c?cccc} 
    \multicolumn{5}{c}{} \\
    & \multicolumn{4}{c}{\textbf{Attaching NoTS to existing architectures}} \\
   \textbf{Dataset}  & PatchTST  & PatchTST + \textbf{NoTS}  & {iTransformer} & iTransformer + \textbf{NoTS} \\
    \hline
  EthanolConcentration & 25.10 & 25.48 & 30.42 & 30.04 \\
    Heartbeat & 73.17 & 74.63 & 73.17 & 74.15 \\
    PEMS-SF & 88.44 & 90.75 & 89.02 & 90.75 \\
    SelfRegulationSCP1 & 78.16 & 82.25 & 87.71 & 88.06 \\
    SelfRegulationSCP2 & 51.67 & 51.67 & 57.78 & 58.33 \\
        \hline
    \textbf{Average} & 63.31 & 64.96 & 67.62 & 68.27 \\
\end{tabular}
  \end{center}
\end{table}

\begin{table}[htbp!]
\setlength{\tabcolsep}{4pt}
\footnotesize
\vspace{-6mm}
  \begin{center}
    \caption{Complete imputation results with masking ratio $12.5\%$ and $25\%$.}
    \label{tab:app_imputation}
    \begin{tabular}{c?cccc?cccc} 
         & \multicolumn{4}{c?}{\textbf{Parameter efficient tuning}} & \multicolumn{4}{c}{\textbf{Full-scale fine-tuning}}  \\
     \textbf{Dataset}  & \textbf{NoTS-lw}  & {Next-pred}  & {bioFAME} & {SimMTM} & \textbf{NoTS-lw} & {Next-pred} & {bioFAME} & {SimMTM} \\ 
        \multicolumn{9}{c}{\textit{12.5\% masking ratio}} \\
        \hline
        ETTm1  & 0.1556 & 0.2832 & 0.1957  & 0.1573 & 0.1194 & 0.1219 & 0.1251 & 0.1207 \\
        ETTm2  & 0.1232 & 0.1774 & 0.1183 & 0.1243 & 0.1110 & 0.1164 & 0.1038 & 0.1041 \\
        ETTh1  & 0.2764  & 0.4569 & 0.2471 & 0.2545 & 0.2091 & 0.2126 & 0.1966 & 0.1947 \\
        ETTh2  & 0.1917 & 0.2692 & 0.1746 & 0.1796 & 0.1615 & 0.1886 & 0.1751 & 0.1632 \\
        \multicolumn{9}{c}{\textit{25\% masking ratio}} \\
        \hline
        ETTm1  & 0.1730   & 0.3280 & 0.2103 & 0.1697 & 0.1246 & 0.1377 & 0.1325 & 0.1244 \\
        ETTm2  & 0.1294  & 0.1789 & 0.1257 & 0.1269 & 0.1205 & 0.1218  & 0.1095 & 0.1093 \\
        ETTh1  & 0.2957  & 0.4738 & 0.2695 & 0.2734 & 0.2266 & 0.2440 & 0.2068 & 0.2078 \\
        ETTh2  & 0.1994 & 0.2707 & 0.1824 & 0.1861 & 0.1653 & 0.1872 & 0.1806 & 0.1681 \\
    \end{tabular}%
    
    \begin{tabular}{c?cccc} 
         \multicolumn{5}{c}{} \\
        & \multicolumn{4}{c}{\textbf{Attaching NoTS to existing architectures}} \\
       \textbf{Dataset}  & PatchTST  & PatchTST + \textbf{NoTS}  & {iTransformer} & iTransformer + \textbf{NoTS} \\
        \multicolumn{5}{c}{\textit{12.5\% masking ratio}} \\
        \hline
        ETTm1  & 0.1791 & 0.1657 & 0.1539 & 0.1662 \\
        ETTm2  & 0.1233 & 0.1193 & 0.1082 & 0.1071 \\
        ETTh1  & 0.3277 & 0.2705 & 0.2325 & 0.2227 \\
        ETTh2  & 0.1817 & 0.1797 & 0.1639 & 0.1609 \\
        \multicolumn{5}{c}{\textit{25\% masking ratio}} \\
        \hline
        ETTm1  & 0.1837 & 0.1903 & 0.1698 & 0.1665 \\
        ETTm2  & 0.1295 & 0.1268 & 0.1140 & 0.1117 \\
        ETTh1  & 0.3668 & 0.2952 & 0.2483 & 0.2318 \\
        ETTh2  & 0.1926 & 0.1827 & 0.1725 & 0.1678 \\
    \end{tabular}%
  \end{center}
\end{table}

In our experiments, iTransformer \citep{liu2023itransformer} is implemented to transform inputs to an embedding dimension of $16$, where we build the decoder to be symmetric and linear. We used a $3$-layer $4$-head transformer network with a token dimensionality of $128$ and $4 \times$ size in the feed forward layer. 
PatchTST \citep{nie2022time} is implemented with a patch length of $16$, stride of $8$, and token dimension of $32$. In cases where PatchTST becomes too computationally heavy (e.g., anomaly detection tasks), we adjust batch size to be $1$ and increase patch length to be $32$.

\subsubsection{Complete experimental results}
\label{app:additional_results}

We show the complete classification results in Table~\ref{tab:app_clf_ucr} and Table~\ref{tab:app_clf_uea} and the complete imputation results in Table~\ref{tab:app_imputation}. The averaged results are presented in Table~\ref{tab:main}.

\paragraph{Imputation task details} We perform a channel-wise imputation task instead of the traditional random imputation task. Specifically, when performing the masking, instead of uniformly sample random elements from $C \times T$ entries of $\rmS \in \mathbb{R}^{C \times T}$, we sample uniformly from $T$ columns and cover inputs from all channels. This is to eliminate the overfitting issues from the data embeddings.

\subsubsection{Additional visualizations}

\begin{figure}[h!]
  \centering
  \includegraphics[width=0.9\textwidth]{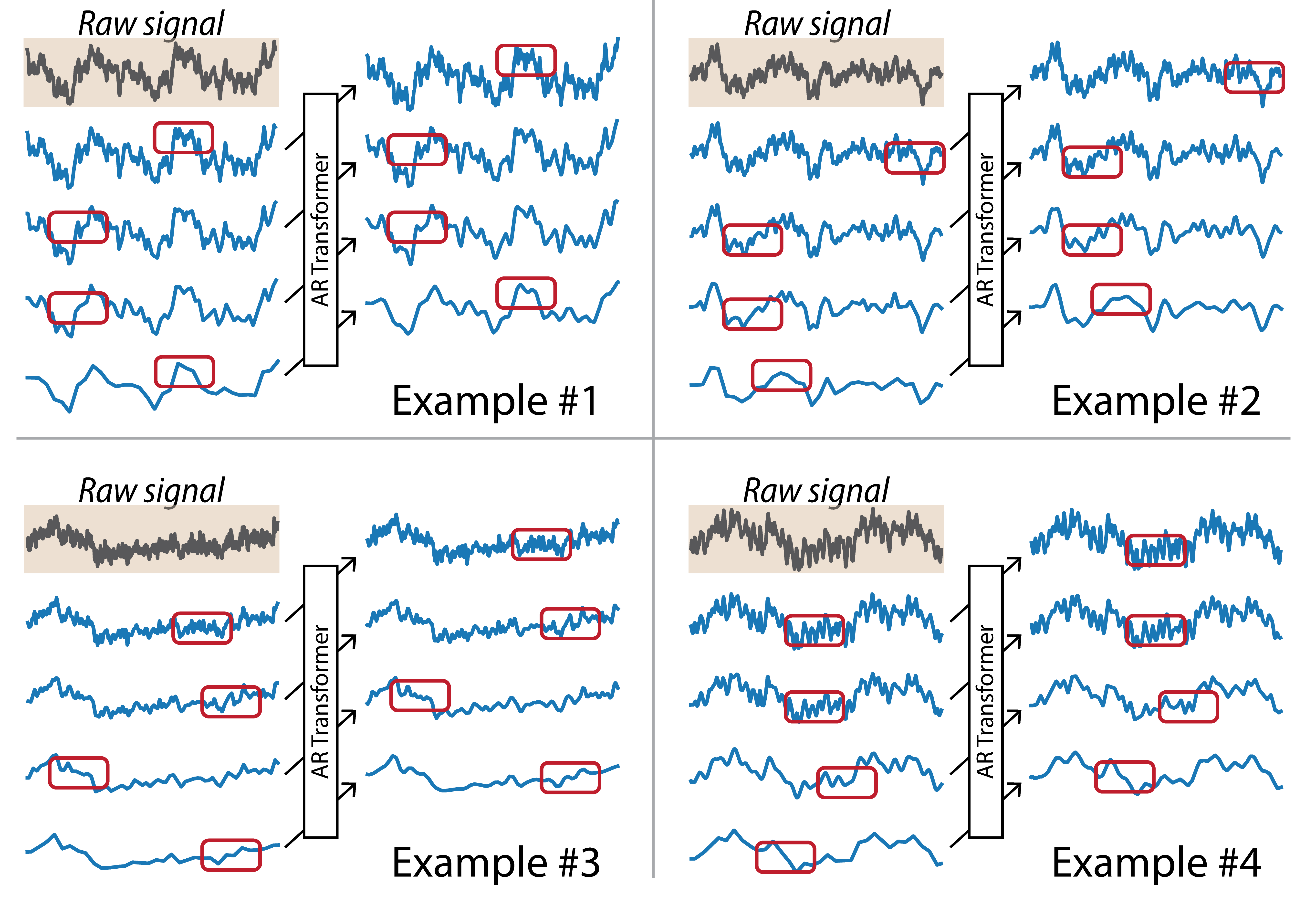}
  \vspace{-3mm}
  \caption{Additional data space visualizations.}
\label{app:data_visualizations_figures}
\end{figure}

\begin{figure}[h!]
  \centering
  \includegraphics[width=\textwidth]{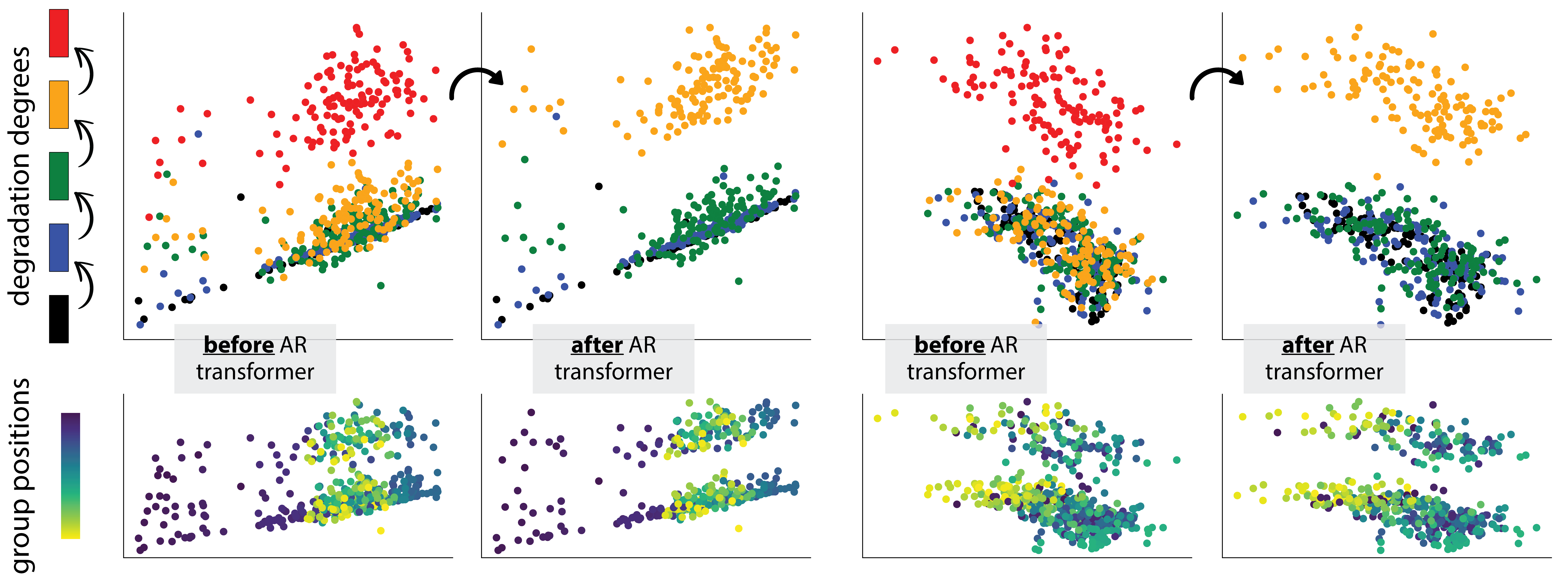}
  \vspace{-3mm}
  \caption{Additional token space visualizations.}
  \label{app:latent_visualizations_figures}
\end{figure}

We present additional data and token space visualizations in Figure~\ref{app:data_visualizations_figures} and Figure~\ref{app:latent_visualizations_figures}, respectively.

\vspace{-3mm}

\end{document}